\pdfoutput=1
\documentclass[sigconf,nonacm]{acmart}

\acmConference[HSCC+ICCPS '27]{30th ACM/IEEE International Conference on Cyber-Physical and Autonomous Systems}{May 17--20, 2027}{Boulder, Colorado, USA}
\acmYear{2027}
\copyrightyear{2027}

\usepackage{algorithm}
\usepackage{algpseudocode}
\usepackage{tikz}
\usetikzlibrary{positioning,arrows.meta,fit,calc,backgrounds,shapes.geometric}
\usepackage{pgfplots}
\pgfplotsset{compat=1.18}
\usepackage{xspace}
\usepackage{booktabs}
\usepackage{xcolor}
\usepackage{enumitem}
\usepackage{tabularx}

\AtEndPreamble{%
  \theoremstyle{acmdefinition}%
  \newtheorem{remark}[theorem]{Remark}%
  \newtheorem{problem}{Problem}}

\newcommand{\tool}{\textsc{BernBRS}\xspace}
\newcommand{\tocheck}[2][]{\textcolor{red}{#2}\ifx\relax#1\relax\else\textcolor{red}{\,[\emph{check: #1}]}\fi}
\newcommand{\R}{\mathbb{R}}
\newcommand{\N}{\mathbb{N}}
\newcommand{\x}{x}
\newcommand{\y}{y}
\newcommand{\z}{z}

\newcommand{\bbf}{b}
\newcommand{\cbf}{c}

\newcommand{\vbf}{v}
\newcommand{\wbf}{w}
\newcommand{\X}{\mathcal{X}}
\newcommand{\Y}{\mathcal{Y}}

\newcommand{\Qcal}{\mathcal{Q}}
\newcommand{\Ical}{\mathcal{I}}
\newcommand{\Ocal}{\mathcal{O}}

\newcommand{\Rcal}{\mathcal{R}}
\newcommand{\Bcal}{\mathcal{B}}

\newcommand{\lo}[1]{\underline{#1}}
\newcommand{\up}[1]{\overline{#1}}
\newcommand{\eps}{\epsilon}
\DeclareMathOperator{\BRS}{BRS}
\DeclareMathOperator{\vol}{vol}

\DeclareMathOperator{\clip}{clip}

\usepackage{caption}

\newcommand{\nummissingtext}[1]{\textcolor{red}{\textlangle\textbf{#1}\textrangle}}
\newcommand{\nummissing}[1]{\ifmmode\text{\nummissingtext{#1}}\else\nummissingtext{#1}\fi}

\makeatletter\@for\armname:=2d,images,vit,tinybert,halfcheetah\do{\IfFileExists{numbers/exp_\armname.tex}{\input{numbers/exp_\armname}}{}}\makeatother
\setcopyright{none}                                  %
\renewcommand\footnotetextcopyrightpermission[1]{}   %

\begin{document}

\title[BernBRS: Certified Backward Reachability for Deep Bernstein Neural Networks]{BernBRS: Certified Backward Reachability \\ for Deep Bernstein Neural Networks}

\author{Ibrahim Albool}
\affiliation{\institution{University of California, Irvine}\city{Irvine}\state{CA}\country{USA}}
\author{Yasser Shoukry}
\affiliation{\institution{University of California, Irvine}\city{Irvine}\state{CA}\country{USA}}

\begin{abstract}
Ensuring the safety of complex neural network systems requires mathematically rigorous verification. While backward reachability analysis is a critical tool for identifying safe initial states, computing tight, formal bounds remains notoriously difficult to scale. Recent advancements have demonstrated that DeepBern-Nets neural networks utilizing Bernstein polynomial activation functions possess advantageous mathematical properties that enable exceptionally fast and tight forward reachability analysis. Building upon this efficiency, this paper introduces \tool, a highly accurate and scalable framework for the backward reachability analysis of DeepBern-Nets. \tool simultaneously computes certified inner and outer approximations of the Backward Reachable Set (BRS). Rather than relying on single convex bounding, our method intelligently refines a cover of the input space by analyzing activation patterns, splitting the input domain precisely where an exact decomposition of the relaxation gap predicts the most significant reduction in error. This targeted refinement enables high levels of scalability to highly complex and massive architectures that remain intractable for existing certified methods. 
We demonstrate the efficacy of our framework on high-dimensional models, including a linearized closed-loop approximation of the Half-Cheetah controller and a distilled Bernstein student of a Transformer block and we empirically study how these model-level guarantees relate to the original Half-Cheetah and TinyBERT classifier, respectively.
\vspace{-2mm}

\end{abstract}

\keywords{Backward reachability, Neural Network Verification, Bernstein polynomials, Preimage Approximation. \vspace{-0mm}}

\maketitle

\section{Introduction}
\label{sec:intro}

Neural networks (NNs) have become foundational components in modern cyber-physical systems (CPS), increasingly serving as complex controllers, advanced perception modules, and learned models of system dynamics. As these systems are deployed in safety-critical environments—from autonomous navigation to industrial control—the demand for rigorous, formal guarantees of their behavior is paramount.

While significant strides have been made in forward verification and reachability analysis~\cite{katz2017reluplex,zhang2018crown,wang2021betacrown,liu2021algorithms,ivanov2019verisig,dutta2019reachability,huang2019reachnn,tran2020nnv,everett2021reachability,schilling2022verification}, understanding a system's complete safety profile requires the ability to reason in reverse. Backward reachability analysis computes the set of all inputs that lead to a specific output region~\cite{mitchell2005time,mitchell2007comparing,bansal2017hamilton}, a capability that is uniquely suited to answering fundamental questions about CPS safety and control. Approximating the backward reachable set (BRS) serves three vital purposes: isolating the initial conditions that lead to catastrophic failures, so that any state outside this set can be certified as safe~\cite{rober2023}, identifying the domain of inputs guaranteed to reach a desired target via inner approximations~\cite{rober2023}, and providing a mechanism to explain network decision-making~\cite{matoba2020exact}. Furthermore, through iterated preimages, this analysis can be extended to verify multi-step behaviors in closed control loops~\cite{vincent2021reachable,rober2022backward}.

Despite its critical utility, backward reachability remains notoriously difficult and has received less attention than forward verification. Neural networks are inherently non-invertible, meaning multiple inputs can map to the same output. Consequently, the true BRS is frequently non-convex, highly disjointed, and topologically complex. Exact backward analysis has largely been restricted to piecewise-affine architectures, such as ReLU networks, using computationally exhaustive techniques like polyhedral marching~\cite{vincent2021reachable} or activation region enumeration~\cite{matoba2020exact}. To manage this complexity, approximation frameworks relax the network with linear bounds. INVPROP~\cite{kotha2023provably} computes outer bounds only, and PREMAP~\cite{premap2025} computes an inner or an outer set in separate runs. The backward reachability method of Rober et al.~\cite{rober2022backward,rober2023}, which uses the linear relaxations of CROWN~\cite{zhang2018crown}, computes outer bounds and keeps one hyper-rectangle per time step, which can be conservative. INVPROP and PREMAP reduce this looseness with unions of convex sets over input splits, and PREMAP also splits on ReLU neurons. Both are formulated for ReLU networks.

Recent advancements have demonstrated that DeepBern-Nets~\cite{khedr2024deepbern} neural network architectures utilizing Bernstein polynomial activation functions offer uniquely advantageous mathematical properties for verification. Specifically, their inherent polynomial structure has been exploited to compute tight forward bounds, in a few seconds even for networks with millions of parameters. Building upon this unprecedented efficiency, this paper asks the question: can we extend these properties to the backward domain? To answer this, we establish a highly accurate and scalable backward reachability framework for DeepBern-Nets. We introduce \tool, a novel algorithmic approach that simultaneously computes certified inner and outer approximations of the Backward Reachable Set (BRS). Our method intelligently refines a cover of the input space by analyzing activation patterns, splitting the input domain precisely where an exact decomposition of the relaxation gap predicts the most significant reduction in error.
The specific contributions of this work are as follows:
\begin{itemize}[leftmargin=*, labelsep=4pt,topsep=0pt]
    \item The \tool Framework: We propose a comprehensive method for computing certified inner and outer approximations of the backward reachable set of DeepBern-Nets~\cite{khedr2024deepbern}, guaranteeing mathematical soundness at every refinement iteration.

    \item Gap-Driven Refinement: We introduce an advanced branching strategy that partitions the input space based on an exact decomposition of the relaxation gap.

    \item Scalability to Complex Architectures: 
    Our framework demonstrates scalability to a diverse range of massive and complex architectures that, to our knowledge, have not previously been handled by certified backward reachability methods. This includes (i) a merged DeepBern-Net representation of a linearized Half-Cheetah closed loop, and (ii) a Bernstein distilled transformer block and classifier of a TinyBERT model. In both cases, \tool provides mathematically rigorous backward reachability bounds for these surrogate networks; the connection to the underlying MuJoCo simulator and to TinyBERT is evaluated empirically rather than formally certified.

\end{itemize}

\section{Problem Setup}
\label{sec:problem}

\subsection{Notation}
\label{sec:notation}
Let $\mathbb{R}$ and $[N]$ denote the set of Real and finite set of positive integers (i.e., $[N] = \{1, \dots, N\}$ for an integer $N \ge 1$), respectively. For a matrix $W$, $W^+ = \max(W, 0)$, $W^- = \min(W, 0)$ and $|W|$ act entrywise, and $W_{j:}$ is the $j$-th row. Vector inequalities hold entrywise. A \emph{box} is $X = [\lo{\x}, \up{\x}] = \prod_{i=1}^{d} [\lo{x}_i, \up{x}_i]$ where $\lo{\x}, \up{\x} \in \R^{d}$, with edge lengths $h_i(X) = \up{x}_i - \lo{x}_i$. For a finite family $\mathcal{S}$ of sets, $\bigcup \mathcal{S}$ is their union. An \emph{interval vector} $[\lo{\z}, \up{\z}]$ \emph{encloses} a map $\varphi$ on a set $S$ if $\varphi(S) \subseteq [\lo{\z}, \up{\z}]$. The symbol $\vol(S)$ is the Lebesgue measure of the set $S$.

\begin{figure}[t]
    \centering
    \definecolor{tolindigo}{HTML}{332288}\definecolor{tolteal}{HTML}{44AA99}
\definecolor{tolsand}{HTML}{DDCC77}\definecolor{tolwine}{HTML}{882255}
\definecolor{tololive}{HTML}{999933}\definecolor{tolgrey}{HTML}{BBBBBB}
\definecolor{toldark}{HTML}{777777}
\begin{tikzpicture}
\begin{axis}[width=0.91\columnwidth,height=0.40\columnwidth,xmin=-2.6,xmax=2.6,ymin=-1.6,ymax=1.35,
 axis lines=none,clip=false,scale only axis,every node/.style={font=\footnotesize}]
\draw[-{Stealth[length=4pt]},thin] (axis cs:-2.6,0)--(axis cs:2.6,0) node[right,inner sep=2pt]{$z$};
\draw[-{Stealth[length=4pt]},thin] (axis cs:-2.6,-1.1)--(axis cs:-2.6,1.35) node[above,inner sep=2pt]{$\sigma(z)$};
\node[above left,inner sep=1.5pt] at (axis cs:-2,0){$\ell$};
\node[below right,inner sep=1.5pt] at (axis cs:2,0){$u$};
\draw[thin] (axis cs:-2,0.04)--(axis cs:-2,-0.04);
\draw[thin] (axis cs:2,0.04)--(axis cs:2,-0.04);
\draw[tololive,dashed,line width=0.6pt] (axis cs:-2,-1)--(axis cs:-0.6667,-0.8)--(axis cs:0.6667,0.6)--(axis cs:2,1);
\fill[tololive] (axis cs:-2,-1) circle(1.6pt);
\fill[tololive] (axis cs:-0.6667,-0.8) circle(1.6pt);
\fill[tololive] (axis cs:0.6667,0.6) circle(1.6pt);
\fill[tololive] (axis cs:2,1) circle(1.6pt);
\node[tololive,below left] at (axis cs:-2,-1){$c_0$};
\node[tololive,below right] at (axis cs:-0.6667,-0.8){$c_1$};
\node[tololive,above left] at (axis cs:0.6667,0.6){$c_2$};
\node[tololive,above left] at (axis cs:2,1){$c_3$};
\addplot[tolindigo,line width=1.1pt,domain=-2:2,samples=60,no markers]
 {-1*(1-(x+2)/4)^3 + (-0.8)*3*((x+2)/4)*(1-(x+2)/4)^2 + 0.6*3*((x+2)/4)^2*(1-(x+2)/4) + ((x+2)/4)^3};
\draw[black!75,densely dashed,line width=0.8pt] (axis cs:-1.3,0.2)--(axis cs:-1.3,-1.3) node[below,black,inner sep=2pt]{$\lo{z}$};
\draw[black!75,densely dashed,line width=0.8pt] (axis cs:1.4,0.9)--(axis cs:1.4,-1.3) node[below,black,inner sep=2pt]{$\up{z}$};
\end{axis}
\end{tikzpicture}
    \caption{A Bernstein activation with its control points and an interval $[\lo{z}, \up{z}]$ inside its bounds.}
    \vspace{-5mm}
    \label{fig:activation}
\end{figure}
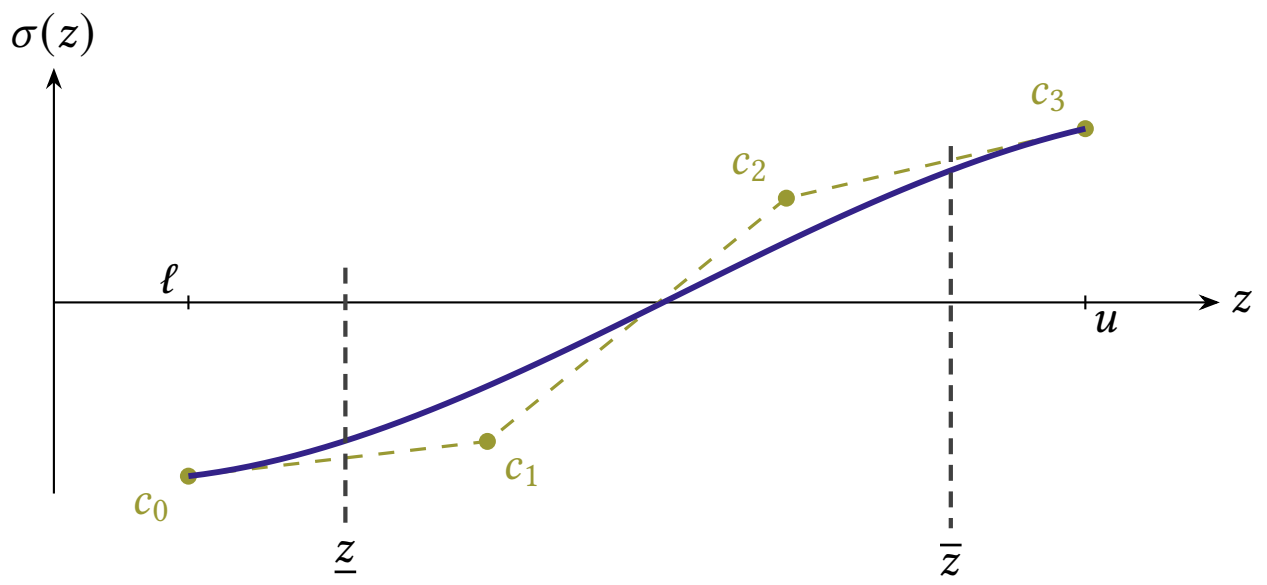

\subsection{Bernstein Polynomials}
\label{sec:bernstein}
In this paper, we focus on neural networks that follow the DeepBern-Nets~\cite{khedr2024deepbern} architecture whose activations are based on the Bernstein polynomials. For an interval $[\ell, u]$ with $\ell < u$, the Bernstein basis polynomials of degree $n$ are:
\begin{equation}
    B^{[\ell,u]}_{n,i}(z) = \binom{n}{i} \frac{(z - \ell)^i\, (u - z)^{n-i}}{(u - \ell)^n}, \qquad i = 0, \dots, n.
    \label{eq:basis}
\end{equation}
A polynomial $q$ of degree at most $n$ has a unique form:
\begin{equation}
q(z) = \sum_{i=0}^{n} c_i\, B^{[\ell,u]}_{n,i}(z).
\end{equation}
The vector $c = (c_0, \dots, c_n)$ holds the \emph{Bernstein coefficients} of $q$ on $[\ell, u]$. We use the following properties of $q$~\cite{khedr2024deepbern,ramshaw1989blossoms,prautzsch2002bezier,farouki2012bernstein}.
\begin{enumerate}
    \item[(P1)] \emph{Enclosure of range}~\cite{farouki2012bernstein}. $\min_i c_i \le q(z) \le \max_i c_i$ for $z \in [\ell, u]$, with $q(\ell) = c_0$ and $q(u) = c_n$.
    \item[(P2)] \emph{Subdivision}~\cite{prautzsch2002bezier}. For $\ell \le \ell' < u' \le u$, the Bernstein coefficients $c'$ of $q$ on $[\ell', u']$ follow from $c$ by de Casteljau steps in $\mathcal{O}(n^2)$ operations, and each $c'_i$ is a convex combination of the entries of $c$. For $\ell' = u'$ we set $c'_i = q(\ell')$ for all $i$.
    \item[(P3)] \emph{Derivative}~\cite{farouki2012bernstein}. $q'(z) = \tfrac{n}{u-\ell} \sum_{i=0}^{n-1} (c_{i+1} - c_i)\, B^{[\ell,u]}_{n-1,i}(z)$.
    \item[(P4)] \emph{Linear precision.} The Bernstein coefficients of degree $n$ of the affine map $z \mapsto \alpha + s z$ on $[\ell, u]$ are $c_i = \alpha + s\,\big(\ell + (u-\ell)\, i/n\big)$ for all $i$.

\end{enumerate}

\subsection{Bernstein Neural Networks}
\label{sec:bernn}

\begin{definition}[Bernstein neural network~\cite{khedr2024deepbern}]
\label{def:bernn}
A Bernstein neural network with $R$ hidden layers of widths $w_r$, input dimension $d$ and output dimension $m$ is the map $f : \R^d \to \R^m$ defined by:
\begin{equation}
\begin{aligned}
    &\z^{(0)}(\x) = \x, \quad \z^{(r)}(\x) = W_r\, \sigma^{(r-1)}\big(\z^{(r-1)}(\x)\big) + \bbf_r, \\
    &f(\x) = W_{R+1}\, \sigma^{(R)}\big(\z^{(R)}(\x)\big) + \bbf_{R+1},
\end{aligned}
\label{eq:bernn}
\end{equation}
for $r \in [R]$, where $\z^{(r)}(\x)$ is the vector of pre-activations of layer $r$, $\sigma^{(0)}$ is the identity and $\sigma^{(r)}$ applies the activation $\sigma^{(r)}_j$ of neuron $(r, j)$ to coordinate $j$. Neuron $(r, j)$ has a degree $n_r$, bounds $[\ell^{(r)}_j, u^{(r)}_j]$ with $\ell^{(r)}_j < u^{(r)}_j$ on its pre-activation, and Bernstein coefficients $c^{(r)}_j \in \R^{n_r+1}$. Its activation is:
\begin{equation}
\begin{aligned}
    \sigma(z; \ell, u, c) &= \sum_{i=0}^{n_r} c_i\, B^{[\ell,u]}_{n_r,i}(z), \quad
    \sigma^{(r)}_j(z) &= \sigma\big(z; \ell^{(r)}_j, u^{(r)}_j, c^{(r)}_j\big).
\end{aligned}
\label{eq:activation}
\end{equation}
\end{definition}

For layer $r$ $W_r \in \R^{w_r \times w_{r-1}}$ and $\bbf_r \in \R^{w_r}$ denote the weight matrix and bias vector of layer $r$, respectively. $d \in \N$ is the input dimension.

\subsection{Activation Patterns}
\label{sec:patterns}

To systematically analyze the network's non-linear behavior, we must first break down its operation into tractable components. We achieve this by introducing \emph{activation patterns}, which constrain the activation functions to specific intervals and allow us to identify the precise subsets of the input domain that lead to those states.
\begin{definition}[Activation pattern]
\label{def:pattern}
An \emph{activation pattern} is defined as $P = (\hat X, \hat Z)$ such that:
\begin{enumerate}
    \item $\hat X \subseteq \R^d$ is a box of network inputs,
    \item $\hat Z = \{\hat Z^{(1)}, \dots, \hat Z^{(R)}\}$. For each layer $r \in [R]$ the interval $\hat Z^{(r)} = [\lo{\vphantom{\up{\hat\z}}\hat\z}^{(r)}, \up{\hat\z}^{(r)}] \subseteq \R^{w_r}$.
\end{enumerate}
Each activation pattern is associated with a \emph{region} $\Rcal(P)$ defined as:
\begin{equation}
    \Rcal(P) = \big\{ \x \in \hat X : \z^{(r)}(\x) \in \hat Z^{(r)} \text{ for all } r \in [R] \big\}.
    \label{eq:region}
\end{equation}
The region $\Rcal(P)$ is non-convex in general.
\end{definition}

\begin{figure*}[t]
\centering
\input{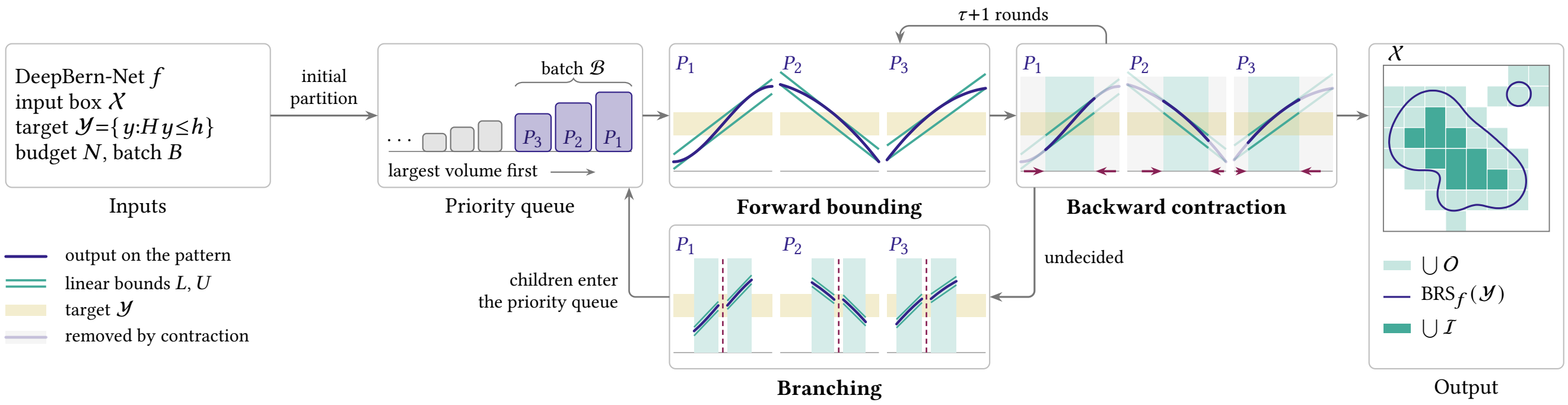}
\vspace{-1mm}
\caption{
Algorithmic Architecture of the \tool Framework. The algorithm begins by partitioning the input domain $\X$ into a priority queue ranked by estimated volume. In each iteration, the largest patterns undergo forward bounding to establish parallel linear bounds ($L, U$ ) enclosing the output. Backward contraction subsequently discards inputs violating the target condition $f(x) \notin \mathcal{Y}$ (faded regions). Forward bounding and backward contraction alternate cyclically for $\tau + 1$ rounds. Undecided patterns branch into refined children and re-enter the queue, while certified regions form the final inner ($\bigcup \Ical$) and outer ($\bigcup \Ocal$) approximations of the target set.
\vspace{-5mm}
}
\label{fig:flow}
\end{figure*}

\subsection{Backward Reachable Sets}
\label{sec:brs}
To define the problem of interest, we assume  an input domain defined as a bounding box $\X \subset \R^d$. We define our target $\Y \subset \R^m$ a polyhedron represented by a system of inequalities:
\begin{equation}
    \Y = \{ \y \in \R^m : H \y \le h \}, \qquad H \in \R^{m_H \times m}, \; h \in \R^{m_H},
    \label{eq:target}
\end{equation}

\begin{definition}[Backward reachable set]
\label{def:brs}
For a given mapping function $f:\X \rightarrow \R^m$ and a target $\Y \subset \R^m$, the backward reachable set (BRS) consists of all inputs points in $\X$ that maps to the target polyhedron $\Y$, that is, $\BRS_f(\Y) = \{ \x \in \X : f(\x) \in \Y \}$.
\end{definition}

We can now state the problem studied in this paper as follows.
\begin{problem}[Certified BRS approximation]
\label{prob:brs}
Given a Bernstein neural network $f$, an input domain $\X$, and a target $\Y$, find finite collections $\Ical$ and $\Ocal$ of subsets of $\X$, with each subset of the form $\Rcal(P) \cap \{\x : A \x \le \cbf\}$, such that
\begin{equation}
    \bigcup \Ical \;\subseteq\; \BRS_f(\Y) \;\subseteq\; \bigcup \Ocal .
    \label{eq:problem}
\end{equation}
\end{problem}

The combined region defined by $\Ical$ forms the \emph{inner} approximation; it provides a strict guarantee that any input $x$ found within it will successfully map to the target ($f(x) \in \mathcal{Y}$). Conversely, the combined region defined by $\Ocal$ forms the \emph{outer} approximation; it guarantees that any input $x \in \mathcal{X}$ falling strictly outside of this combined region will fail to reach the target ($f(x) \notin \mathcal{Y}$).

\section{Overview of the \tool Framework}
\label{sec:overview}

As shown in Figure~\ref{fig:flow} and Algorithms~\ref{alg:overview} and~\ref{alg:pattern}, \tool computes certified inner and outer approximations of the BRS of a Bernstein neural network. Its inputs are the network $f$, the input box $\X$, the polyhedral target $\Y = \{y : H y \le h\}$ and a budget of $N$ patterns (or partitions). Its outputs are two finite collections  of sets $\Ical$ and $\Ocal$ with $\bigcup \Ical \subseteq \BRS_f(\Y) \subseteq \bigcup \Ocal$.
The overarching architecture of this approach relies on a continuous, iterative cycle of bounding, contraction, and branching, initialized by establishing a root pattern that encapsulates the entire input domain $\X$ and inserting it into a volumetrically prioritized queue, $\Qcal$ (Algorithm~\ref{alg:overview}, line~1). Algorithm~\ref{alg:overview} gives the loop over this queue, and Algorithm~\ref{alg:pattern} gives the processing of one pattern.

Fundamental to the \tool framework is the partitioning of the input space via activation patterns, which explicitly couple an input spatial box with a specific pre-activation interval for every neuron within the network architecture. To evaluate these patterns, the system employs forward bounding (Algorithm~\ref{alg:pattern}, line~2), sequentially computing limits from the input layer through to the final output layer. This is achieved through interval bound propagation, wherein the post-activation intervals of one layer dictate the pre-activation intervals of the subsequent layer via affine mapping. During this forward pass, each non-linear activation function is strictly enclosed by parallel upper and lower affine bounds. The vertical distance separating these parallel lines defines the relaxation width for each individual neuron. By back-substituting these relaxations through the network layers, the algorithm ultimately generates linear lower bounds, $L_i(x)$, and upper bounds, $U_i(x)$, over the target rows for all inputs contained within the specified pattern.

Following the forward pass, the framework implements a backward contraction pass (Algorithm~\ref{alg:pattern}, line~7) designed to systematically identify and prune input sub-regions that demonstrably fail to map into the polyhedral target. This contraction operates in reverse, propagating constraints from the network output back toward the input domain. Initially, the defined target boundaries and the output layer weights are utilized to narrow the post-activation intervals of the final hidden layer. Subsequently, neurons that are mathematically certified as monotone undergo an inversion process. This mathematical inversion allows the tighter post-activation limits to iteratively constrain and shrink their corresponding pre-activation intervals. To maximize the reduction of the pattern's geometric region, this contraction pass alternates with forward bounding for up to $\tau+1$ rounds (Algorithm~\ref{alg:pattern}, lines~1--12).

The algorithm continuously evaluates its progress toward definitive classification by measuring the gap between the computed bounds, mathematically defined as $\delta_i = U_i(x) - L_i(x)$. A target row achieves certified satisfaction if its corresponding upper bound remains strictly less than or equal to the defined target threshold, expressed as $U_i(x) \le h_i$ for all $x$ in the pattern (Algorithm~\ref{alg:pattern}, line~5). Conversely, if the lower bound exceeds this threshold ($L_i(x) > h_i$ for all $x$ in the pattern), the row is certified as violated, and the corresponding spatial region is entirely pruned from the search space. Patterns where the bounds cross the threshold are designated as undecided. For these undecided regions, the verified inner subsets are dynamically allocated to the collection $\Ical$, while their enveloping outer subsets are concurrently allocated to $\Ocal$ (Algorithm~\ref{alg:pattern}, lines~13 and~19).

To resolve undecided patterns and refine the computational bounds, the framework employs a gap-driven branching strategy (Algorithm~\ref{alg:pattern}, line~15) that partitions these regions into smaller child patterns. A naive approach of bisecting every interval would induce an intractable exponential explosion. For instance, a $400$-neuron network would theoretically generate up to $2^{400}$ child patterns. To ensure computational scalability, \tool computes reduction scores for both input dimensions and individual neurons to identify the specific split that is expected to reduce the target gap the most. If an input dimension yields the highest score, a standard spatial split is executed, yielding two distinct child patterns. Alternatively, if neurons achieve higher reduction scores, the system executes a ``joint split'' across a selected subset of neurons. This joint splitting mechanism utilizes the means and standard deviations of uniform samples to tightly confine the selected neurons to a narrow central interval. By strategically nesting these restrictions, splitting $p$ neurons generates at most $2p+1$ children, vastly outperforming the exponential $3^p$ growth that would otherwise occur without such statistical confinement.

The orchestration of this entire verification procedure is governed by a priority queue that continuously ranks pending activation patterns based on their estimated volumetric size relative to the original input bounding box $\X$. During each iterative round, the algorithm extracts a batch consisting of up to $B$ patterns that exhibit the largest estimated volumes (Algorithm~\ref{alg:overview}, line~3) and processes them in parallel with Algorithm~\ref{alg:pattern} (Algorithm~\ref{alg:overview}, lines~5--7). The computational process officially terminates either when the priority queue $\Qcal$ is completely emptied or when the pre-allocated pattern budget $N$ is fully exhausted (Algorithm~\ref{alg:overview}, line~2). Upon termination, any patterns that remain unresolved within the queue are subjected to a final classification pass (Algorithm~\ref{alg:overview}, lines~10--14), after which their respective geometric regions are formally finalized and sorted into the ultimate output collections $\Ical$ and $\Ocal$. We provide technical details of each of the three components (bounding, contraction, and branching) in the next sections.

\begin{algorithm}[t]
\caption{\tool: parallel refinement over activation patterns}
\label{alg:overview}
\begin{algorithmic}[1]
\Require Network $f$, input box $\X$, target $\Y = \{H\y \le h\}$, budget $N$, batch size $B$, volume gate $\gamma \in (0,1]$
\Ensure Collections $\Ical$, $\Ocal$ satisfying~\eqref{eq:problem}
\State $\Qcal \gets$ the root pattern, or the initial partition; %
\; $\Ical, \Ocal \gets \emptyset$; \; $\nu \gets 0$
\While{$\Qcal \neq \emptyset$ \textbf{and} $\nu < N$} \Comment{refinement rounds}
    \State $\Bcal \gets$ the patterns of $\Qcal$ of largest estimated volume, at most $\min(B, N - \nu)$, each of estimated volume at least $\gamma$ times the largest
    \State $\Qcal \gets \Qcal \setminus \Bcal$; \; $\nu \gets \nu + |\Bcal|$
    \For{\textbf{each} $P \in \Bcal$ \textbf{in parallel}}
        \State $\mathit{res}_P \gets \textsc{ProcessPattern}(P, \textsc{true})$ \Comment{Alg.~\ref{alg:pattern}}
    \EndFor
    \State \textsc{Collect}$(\{\mathit{res}_P\})$
\EndWhile
\State $\Qcal' \gets \Qcal$, or the $\min(N, |\Qcal|)$ patterns of $\Qcal$ of largest estimated volume \Comment{final classification}
\For{\textbf{each} $P \in \Qcal'$ \textbf{in parallel}}
    \State $\mathit{res}_P \gets \textsc{ProcessPattern}(P, \textsc{false})$
\EndFor
\State \textsc{Collect}$(\{\mathit{res}_P\})$; \; $\Ocal \gets \Ocal \cup \{\Rcal(P) : P \in \Qcal \setminus \Qcal'\}$
\State \Return $(\Ical, \Ocal)$
\end{algorithmic}
\end{algorithm}

\begin{algorithm}[t]
\caption{$\textsc{ProcessPattern}(P, s)$}
\label{alg:pattern}
\begin{algorithmic}[1]
\Require Pattern $P = (\hat X, \hat Z)$, splitting flag $s$, contraction rounds $\tau$, depth limit $D$
\For{$t = 0, \dots, \tau$}
    \State $\Bcal_P, (L, U) \gets$ certified bounds over $\Rcal(P)$ \Comment{bounding, Alg.~\ref{alg:bounds}}
    \If{$\Bcal_P$ is empty \textbf{or} some row is certified violated} \Return \textsc{prune}
    \EndIf
    \If{every row is certified satisfied} \Return $\textsc{accept}(\Rcal(P))$
    \EndIf
    \State $P' \gets$ contraction of $P$ against $\Y$ \Comment{Alg.~\ref{alg:tighten}}
    \If{$P'$ is empty} \Return \textsc{prune}
    \EndIf
    \If{$t < \tau$} $P \gets P'$
    \EndIf
\EndFor
\State $I \gets \Rcal(P') \cap \{ U(\x) \le h \}$, \; $O \gets \Rcal(P') \cap \{ L(\x) \le h \}$
\If{$s$ \textbf{and} $\mathrm{depth}(P) < D$}
    \State $\mathcal{C} \gets \textsc{Refine}(P', L, U)$ \Comment{branching, Alg.~\ref{alg:refine}}
    \If{$\mathcal{C} \neq \emptyset$} \Return $\textsc{split}(\mathcal{C})$
    \EndIf
\EndIf
\State \Return $\textsc{undecided}(I, O)$
\end{algorithmic}
\end{algorithm}

\section{Forward Bounding}
\label{sec:bounds}

This section explains the mathematical steps behind the forward bounding method. Forward bounding is a crucial step used to safely estimate the range of possible outputs of each neuron. We provide the complete, step-by-step process in Algorithm~\ref{alg:bounds}. All mathematical proofs  are included in Appendix~\ref{app:proofs:phase1}.

\subsection{Interval Propagation over a Pattern}
\label{sec:ibp}
For every activation pattern $P = (\hat{X}, \hat{Z})$ prioritized within the queue $\Qcal$, the procedure starts by applying the interval bound propagation from ~\cite{khedr2024deepbern}, restricted to the pattern. This propagation starts from $[\lo{\sigma}^{(0)}, \up{\sigma}^{(0)}] = \hat X$ and proceeds iteratively to derive rigorous bounds for each subsequent layer $r\in[R]$. The bounds are computed as:
\begin{align}
    \lo{\z}^{(r)} &= \max\big( W_r^+ \lo{\sigma}^{(r-1)} + W_r^- \up{\sigma}^{(r-1)} + \bbf_r, \; \lo{\hat\z}^{(r)} \big), \nonumber\\
    \up{\z}^{(r)} &= \min\big( W_r^+ \up{\sigma}^{(r-1)} + W_r^- \lo{\sigma}^{(r-1)} + \bbf_r, \; \up{\hat\z}^{(r)} \big), \label{eq:ibp_affine}\\
    [\lo{\sigma}^{(r)}_j, \up{\sigma}^{(r)}_j] &= \big[\min_i c'_i, \max_i c'_i\big]. \label{eq:ibp_act}
\end{align}
The specific Bernstein coefficients $c'$ are systematically derived by leveraging the subdivision property (Property P2), which mathematically restricts the polynomial activation domain for each individual neuron  to the interval $[\lo{z}^{(r)}_j, \up{z}^{(r)}_j]$.

\subsection{Linear Relaxations of Bernstein Activations}
\label{sec:relaxation}
The subsequent step necessitates the creation of certified linear over- and under-approximations for every Bernstein activation function. For an isolated interval $[\underline{z},\overline{z}]$, the primary secant line intersecting the initial and final control points of the activation curve defines a shared slope for both bounding lines. By leveraging the intrinsic mathematical properties of the Bernstein coefficients, the precise vertical offsets required to guarantee bounding validity can be computed algebraically, thereby circumventing the computational overhead of nonlinear curve searches. This is captured by the following result.

\begin{lemma}[Bernstein linear relaxation]
\label{lem:relaxation}
Let $c'$ be the Bernstein coefficients of $\sigma^{(r)}_j$ on $[\lo{z}, \up{z}]$, and let $s \in \R$ be any slope. Define:
\begin{align*}
    e_i &= c'_i - s\,\xi_i, \quad \xi_i = \lo{z} + (\up{z} - \lo{z})\, i / n, \quad i = 0, \dots, n,
\end{align*}
and let $\lo\beta = \min_i e_i$ and $\up\beta = \max_i e_i$. Then:
\begin{equation}
    s\, z + \lo\beta \;\le\; \sigma^{(r)}_j(z) \;\le\; s\, z + \up\beta \qquad \text{for all } z \in [\lo{z}, \up{z}].
    \label{eq:relaxation}
\end{equation}
\end{lemma}

According to the linear precision property (Property P4) of Bernstein polynomials, any affine mapping $z\mapsto sz$ on the interval $[\underline{z},\overline{z}]$ is exactly represented by $n+1$ Bernstein coefficients $s\xi_{i}$. Consequently, the term $e_{i}$ quantifies the vertical displacement of the $i$-th control point of $\sigma_{j}^{(r)}$ relative to this linear approximation, effectively serving as the $i$-th Bernstein coefficient of the residual function $\sigma_{j}^{(r)}(z)-sz$. The enclosure property (Property P1) ensures that the discrete extrema of these offsets strictly bound the continuous residual function over the defined interval, establishing mathematically sound offsets for the parallel relaxation lines. The \tool architecture utilizes the exact secant slope $s=(c_{n}^{\prime}-c_{0}^{\prime})/(\overline{z}-\underline{z})$, which yields a total relaxation width defined by $\beta=\overline{\beta}-\underline{\beta}$.

\begin{figure}[t]
    \centering
    \definecolor{tolindigo}{HTML}{332288}
\definecolor{tolteal}{HTML}{44AA99}
\definecolor{tololive}{HTML}{999933}
\definecolor{tolwine}{HTML}{882255}
\definecolor{toldark}{HTML}{777777}
\begin{tikzpicture}[x=3.3cm,y=0.3326cm,>=Stealth]
\begin{scope}[shift={(0.00,0)}]
\draw[toldark,line width=0.3pt] (0,-1.600) -- (1,-1.600);
\node[toldark,font=\scriptsize,anchor=north] at (0,-1.600) {$\ell'$};
\node[toldark,font=\scriptsize,anchor=north] at (1,-1.600) {$u'$};
\node[font=\scriptsize,anchor=south west,inner sep=1pt] at (0,2.350) {$J=1$};
\draw[tolteal,line width=0.8pt] (0,-1.480) -- (1,-0.880);
\draw[tolteal,line width=0.8pt] (0,2.080) -- (1,2.680);
\draw[tololive,dashed,line width=0.5pt] (0.000,0.000) -- (0.200,2.200) -- (0.400,-1.200) -- (0.600,2.000) -- (0.800,-1.000) -- (1.000,0.600);
\fill[tololive] (0.000,0.000) circle (1.1pt);
\fill[tololive] (0.200,2.200) circle (1.1pt);
\fill[tololive] (0.400,-1.200) circle (1.1pt);
\fill[tololive] (0.600,2.000) circle (1.1pt);
\fill[tololive] (0.800,-1.000) circle (1.1pt);
\fill[tololive] (1.000,0.600) circle (1.1pt);
\draw[tolindigo,line width=1.0pt] plot[smooth] coordinates {(0.000,0.000) (0.017,0.168) (0.033,0.309) (0.050,0.424) (0.067,0.518) (0.083,0.593) (0.100,0.650) (0.117,0.693) (0.133,0.723) (0.150,0.742) (0.167,0.752) (0.183,0.755) (0.200,0.752) (0.217,0.743) (0.233,0.731) (0.250,0.715) (0.267,0.698) (0.283,0.679) (0.300,0.660) (0.317,0.640) (0.333,0.620) (0.350,0.600) (0.367,0.581) (0.383,0.563) (0.400,0.546) (0.417,0.529) (0.433,0.513) (0.450,0.498) (0.467,0.484) (0.483,0.470) (0.500,0.456) (0.517,0.443) (0.533,0.429) (0.550,0.415) (0.567,0.401) (0.583,0.387) (0.600,0.371) (0.617,0.355) (0.633,0.338) (0.650,0.320) (0.667,0.301) (0.683,0.282) (0.700,0.262) (0.717,0.241) (0.733,0.221) (0.750,0.201) (0.767,0.182) (0.783,0.164) (0.800,0.149) (0.817,0.137) (0.833,0.129) (0.850,0.127) (0.867,0.130) (0.883,0.142) (0.900,0.163) (0.917,0.196) (0.933,0.241) (0.950,0.302) (0.967,0.381) (0.983,0.479) (1.000,0.600)};
\draw[toldark,line width=0.4pt,{Stealth[length=2pt]}-{Stealth[length=2pt]}] (1.07,-0.880) -- (1.07,2.680);
\node[font=\scriptsize,anchor=west,inner sep=1.5pt] at (1.07,0.900) {$\beta$};
\end{scope}
\begin{scope}[shift={(1.27,0)}]
\draw[toldark,line width=0.3pt] (0,-1.600) -- (1,-1.600);
\node[toldark,font=\scriptsize,anchor=north] at (0,-1.600) {$\ell'$};
\node[toldark,font=\scriptsize,anchor=north] at (1,-1.600) {$u'$};
\node[font=\scriptsize,anchor=south west,inner sep=1pt] at (0,2.350) {$J=2$};
\draw[tolwine,dashed,line width=0.5pt] (0.5,-1.600) -- (0.5,2.680);
\draw[tolteal,line width=0.8pt] (0,-0.740) -- (1,-0.140);
\draw[tolteal,line width=0.8pt] (0,1.040) -- (1,1.640);
\draw[tololive,dashed,line width=0.5pt] (0.000,0.000) -- (0.100,1.100) -- (0.200,0.800) -- (0.300,0.625) -- (0.400,0.538) -- (0.500,0.456);
\fill[tololive] (0.000,0.000) circle (1.1pt);
\fill[tololive] (0.100,1.100) circle (1.1pt);
\fill[tololive] (0.200,0.800) circle (1.1pt);
\fill[tololive] (0.300,0.625) circle (1.1pt);
\fill[tololive] (0.400,0.538) circle (1.1pt);
\fill[tololive] (0.500,0.456) circle (1.1pt);
\draw[tololive,dashed,line width=0.5pt] (0.500,0.456) -- (0.600,0.375) -- (0.700,0.300) -- (0.800,0.150) -- (0.900,-0.200) -- (1.000,0.600);
\fill[tololive] (0.500,0.456) circle (1.1pt);
\fill[tololive] (0.600,0.375) circle (1.1pt);
\fill[tololive] (0.700,0.300) circle (1.1pt);
\fill[tololive] (0.800,0.150) circle (1.1pt);
\fill[tololive] (0.900,-0.200) circle (1.1pt);
\fill[tololive] (1.000,0.600) circle (1.1pt);
\draw[tolindigo,line width=1.0pt] plot[smooth] coordinates {(0.000,0.000) (0.017,0.168) (0.033,0.309) (0.050,0.424) (0.067,0.518) (0.083,0.593) (0.100,0.650) (0.117,0.693) (0.133,0.723) (0.150,0.742) (0.167,0.752) (0.183,0.755) (0.200,0.752) (0.217,0.743) (0.233,0.731) (0.250,0.715) (0.267,0.698) (0.283,0.679) (0.300,0.660) (0.317,0.640) (0.333,0.620) (0.350,0.600) (0.367,0.581) (0.383,0.563) (0.400,0.546) (0.417,0.529) (0.433,0.513) (0.450,0.498) (0.467,0.484) (0.483,0.470) (0.500,0.456) (0.517,0.443) (0.533,0.429) (0.550,0.415) (0.567,0.401) (0.583,0.387) (0.600,0.371) (0.617,0.355) (0.633,0.338) (0.650,0.320) (0.667,0.301) (0.683,0.282) (0.700,0.262) (0.717,0.241) (0.733,0.221) (0.750,0.201) (0.767,0.182) (0.783,0.164) (0.800,0.149) (0.817,0.137) (0.833,0.129) (0.850,0.127) (0.867,0.130) (0.883,0.142) (0.900,0.163) (0.917,0.196) (0.933,0.241) (0.950,0.302) (0.967,0.381) (0.983,0.479) (1.000,0.600)};
\draw[toldark,line width=0.4pt,{Stealth[length=2pt]}-{Stealth[length=2pt]}] (1.07,-0.140) -- (1.07,1.640);
\node[font=\scriptsize,anchor=west,inner sep=1.5pt] at (1.07,0.750) {$\beta_J$};
\end{scope}
\end{tikzpicture}
    \vspace{-1mm}
    \caption{Relaxation of a non-monotone activation of degree 5 with the secant slope. Left: offsets from the control points on $[\lo{z}, \up{z}]$ ($J = 1$). Right: the same slope with offsets from two sub-intervals ($J = 2$), which gives a smaller width.}
    \label{fig:relaxJ}
    \vspace{-3mm}
\end{figure}
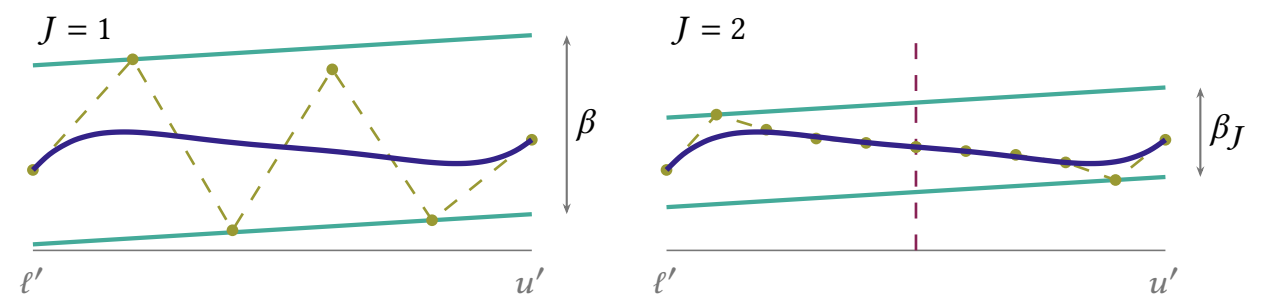

While standard Bernstein coefficients reliably enclose non-monotone activations, they may occasionally induce overly conservative geometric bounds (as shown in Figure~\ref{fig:relaxJ}). To mitigate this, \tool partitions the input domain into $J=2$ equal sub-intervals, extracting bounding offsets from their localized coefficients to formulate significantly tighter enclosures. Although this localized refinement imposes $J$ additional restrictions per neuron, these constraints cascade through the linear bounds to progressively tighten the reachable intervals of subsequent, deeper layers. Ultimately, the maximum width of this relaxation is strictly upper-bounded by a constant factor multiplied by the squared interval width (Proposition~\ref{prop:band}).

\subsection{Linear Bounds on the Target Rows}
\label{sec:linear}

Akin to the linear bound propagation used by tools like CROWN~\cite{zhang2018crown,shi2025genbab}, we map the linear relaxations localized at each neuron to compute the final network-wise linear approximation. To that end, the neuron-wise linear bounds are systematically composed via back-substitution—propagating from the target polyhedral constraints down to the input space—adapting the established principles of linear bound propagation. This is captured by the following theorem.

\begin{theorem}[Linear bounds and gap identity]
\label{thm:linear}
Let every neuron $j$ of every layer $r$ be relaxed on an interval that contains $z^{(r)}_j(\x)$ for all $\x \in \Rcal(P)$. Let $s^{(r)}$, $\lo\beta^{(r)}$ and $\up\beta{}^{(r)}$ be the vectors of slopes, lower offsets and upper offsets of layer $r$. Define $\Lambda^{(R)} = H W_{R+1}$, $o_{\mathrm{out}} = H \bbf_{R+1}$, and for $r = R, \dots, 1$:
\begin{equation}
    \Gamma^{(r)} = \Lambda^{(r)} \operatorname{diag}(s^{(r)}), \qquad \Lambda^{(r-1)} = \Gamma^{(r)} W_r .
\end{equation}
Let $A = \Lambda^{(0)} \in \R^{m_H \times d}$, and let $\Lambda^{(r)+} = \max(\Lambda^{(r)}, 0)$ and $\Lambda^{(r)-} = \min(\Lambda^{(r)}, 0)$ entry-wise. Define
\begin{align}
    \cbf_{\mathrm{lo}} &= o_{\mathrm{out}} + \sum_{r=1}^R \Big( \Lambda^{(r)+} \lo\beta^{(r)} + \Lambda^{(r)-} \up\beta{}^{(r)} + \Gamma^{(r)} \bbf_r \Big), \\
    \cbf_{\mathrm{up}} &= o_{\mathrm{out}} + \sum_{r=1}^R \Big( \Lambda^{(r)+} \up\beta{}^{(r)} + \Lambda^{(r)-} \lo\beta^{(r)} + \Gamma^{(r)} \bbf_r \Big),
\end{align}
and the linear bounds $L(\x) = A\x + \cbf_{\mathrm{lo}}$ and $U(\x) = A\x + \cbf_{\mathrm{up}}$. Then for every $\x \in \Rcal(P)$:
\begin{equation}
    L(\x) \le H f(\x) \le U(\x) \quad \text{and} \quad U(\x) - L(\x) = \sum_{r=1}^R |\Lambda^{(r)}| \, \beta^{(r)},
    \label{eq:gap_identity}
\end{equation}
where $|\Lambda^{(r)}|$ is the entry-wise absolute value and $\beta^{(r)} = \up\beta{}^{(r)} - \lo\beta^{(r)}$ is the vector of relaxation widths of layer $r$. The \emph{gap of row $i$} is the $i$-th entry $\delta_i$ of $U(\x) - L(\x)$.
\end{theorem}

While Theorem~\ref{thm:linear} formulates continuous bounding functions of the input state $x$, definitively resolving a discrete activation pattern necessitates absolute numerical upper and lower bounds for each target row evaluated over the region $\mathcal{R}(P)$. The recursive back-substitution process yields partial linear bounds parameterized in terms of intermediate states, $\sigma^{(r)}(z^{(r)})$ or $z^{(r)}$, at every intermediate layer, culminating in the explicit input-space bounds $L$ and $U$. Because the extreme values of each partial layer-wise bound remain mathematically valid across the entire region $\mathcal{R}(P)$, \tool systematically isolates and retains the tightest observed extrema across all computational layers. These optimally tight extrema constitute the final row bounds $[\underline{\rho}, \overline{\rho}]$ , which are ultimately responsible for either mathematically pruning or validating the activation pattern. The integration of intermediate deep-layer constraints is critical, as split limits imposed on deeper neurons systematically tighten the bounds solely via localized interval restrictions. By recursively applying this identical bounding procedure to the network weights $W_r$ and biases $b_r$ for layers $r \ge 2$, the algorithm tightens the localized pre-activation intervals $[\underline{z}^{(r)}, \overline{z}^{(r)}]$  immediately prior to applying the linear relaxation at layer $r$.

\begin{algorithm}[t]
\caption{Certified bounds over $\Rcal(P)$ (forward bounding)}
\label{alg:bounds}
\begin{algorithmic}[1]
\Require Pattern $P = (\hat X, \hat Z)$
\Ensure Interval bounds $\Bcal_P$, linear bounds $(A, \cbf_{\mathrm{lo}}, \cbf_{\mathrm{up}})$, row bounds $[\lo\rho, \up\rho]$, coefficients $\Lambda^{(r)}$, relaxation widths $\beta^{(r)}$
\State Interval bounds of every layer by Lemma~\ref{lem:ibp}; \textbf{if} some interval is empty \Return \textsc{empty}
\For{$r = 1, \dots, R$}
    \If{$r \ge 2$} intersect $[\lo{\z}^{(r)}, \up{\z}^{(r)}]$ with the bounds from the rows of $W_r$; recompute $[\lo{\sigma}^{(r)}, \up{\sigma}^{(r)}]$ and the certificates
    \EndIf
    \State Relax every neuron of layer $r$ (Lemma~\ref{lem:relaxation}, secant slope)
\EndFor
\State Back-substitute the target rows (Theorem~\ref{thm:linear}); compute the row bounds
\State \Return all bounds
\end{algorithmic}
\end{algorithm}

\section{Backward Contraction}
\label{sec:tightening}

This section formalizes the theoretical foundations of the backward contraction methodology introduced in Section~\ref{sec:overview} with the corresponding proofs provided in Appendix~\ref{app:proofs:phase2}.

\subsection{Backward Contraction over One Layer}

Our algorithm executes the interval bound propagation in reverse across a single network layer. This operation requires the forward bounds $[\underline{\sigma},\overline{\sigma}]$ of the incoming post-activations and the backward bounds $[\underline{z}^{\prime},\overline{z}^{\prime}]$ of the outgoing pre-activations as inputs to yield the refined backward bounds $[\underline{\sigma}^{\prime},\overline{\sigma}^{\prime}]$ of the entering post-activations. For any hidden layer $r$, the affine mapping is defined as $z=W\sigma+b$, where $W=W_{r}$, $b=b_{r}$, and $\sigma=\sigma^{(r-1)}(z^{(r-1)})$. At the output layer, the target constraint $HW_{R+1}\sigma^{(R)}(z^{(R)})+Hb_{R+1}\le h$ necessitates substituting $W=HW_{R+1}$, $b=Hb_{R+1}$, and $\sigma=\sigma^{(R)}(z^{(R)})$. Consequently, the output backward bounds are constrained solely by upper limits defined by $h$, completely lack lower limits, and rely entirely on exact network weights rather than geometric relaxations. The next lemma captures these computations and establishes the correctness of our approach.

\begin{lemma}[Backward interval propagation over one layer]
\label{lem:rows}
Let $\sigma \in [\lo{\sigma}, \up{\sigma}]$ and $\z = W \sigma + \bbf \in [\lo{\z}', \up{\z}']$. For every row $j$ and every $k$ with $W_{jk} \neq 0$,
\begin{align*}
    W_{jk}\, \sigma_k \in \Big[\, & \lo{z}'_j - b_j - \sum_{k' \neq k} \big( W^+_{jk'}\, \up{\sigma}_{k'} + W^-_{jk'}\, \lo{\sigma}_{k'} \big), \\
    & \up{z}'_j - b_j - \sum_{k' \neq k} \big( W^+_{jk'}\, \lo{\sigma}_{k'} + W^-_{jk'}\, \up{\sigma}_{k'} \big) \Big],
\end{align*}
where the two sums are the largest and the smallest value of the other terms of row $j$, as in~\eqref{eq:ibp_affine}.

\end{lemma}

By isolating $\sigma_{k}$ and dividing by the weight $W_{jk}$, one obtains an updated operational interval for $\sigma_{k}$, wherein the interval endpoints are strictly inverted if $W_{jk}<0$. The final backward bound $[\underline{\sigma}_{k}^{\prime},\overline{\sigma}_{k}^{\prime}]$ is established by intersecting these row-wise intervals across all rows $j$ with the existing forward bound $[\underline{\sigma}_{k},\overline{\sigma}_{k}]$, ensuring that the true activation $\sigma_{k}$ is rigorously confined within this intersection. If a bound in $[\underline{z}_{j}^{\prime},\overline{z}_{j}^{\prime}]$ lacks a finite endpoint, it mathematically imposes no constraint from that corresponding direction. Furthermore, an empty intersection definitively proves that no feasible $\sigma$ exists to satisfy the given constraints.

\subsection{Backward Contraction for Monotone Activation Patterns}
For the special case when the activation function is certified monotone (Lemma~\ref{lem:activation} provides the sufficient conditions to check if an activation is monotone), one can compute tighter bounds by exploiting such monotonicity as follows. The newly computed backward bound $[\underline{\sigma}^{\prime},\overline{\sigma}^{\prime}]$ of a post-activation is compared directly against its corresponding forward bound. Endpoints that fail to provide tighter constraints leave the corresponding pre-activation boundaries entirely unaltered. Conversely, a tighter endpoint signifies a function crossing strictly within the domain $[\underline{z},\overline{z}]$, which the BERNBRS framework dynamically isolates using a fixed-iteration bisection method (configured to 60 halvings in empirical evaluations). By retaining the outer boundary of the terminal bisection bracket, the algorithm guarantees that the contracted pre-activation bound $[\underline{z}^{\prime},\overline{z}^{\prime}]$ remains structurally conservative—potentially wider than the exact pre-image, but strictly never narrower as captured by the following result.

\begin{lemma}[Inverse of a certified monotone activation]
\label{lem:inverse}
Let $\sigma = \sigma^{(r)}_j$ be certified non-decreasing on the interval $[\lo{z}, \up{z}]$ (item~2 of Lemma~\ref{lem:clamped}), and let $[\lo{\sigma}', \up{\sigma}']$ be the allowed post-activation interval. Define the \emph{contracted inverse interval} $[\lo{z}', \up{z}']$ as follows.
\begin{enumerate}
    \item \emph{Empty.} If $\sigma(\up{z}) < \lo{\sigma}'$ or $\sigma(\lo{z}) > \up{\sigma}'$, the output is empty.
    \item \emph{Lower end.} If $\sigma(\lo{z}) \ge \lo{\sigma}'$, then $\lo{z}' = \lo{z}$. Otherwise, start from $[a, b] = [\lo{z}, \up{z}]$, for which $\sigma(a) < \lo{\sigma}' \le \sigma(b)$. Replace $[a, b]$ a fixed number of times by the half that keeps this property. Then $\lo{z}' = a$.
    \item \emph{Upper end.} If $\sigma(\up{z}) \le \up{\sigma}'$, then $\up{z}' = \up{z}$. Otherwise, halve $[\lo{z}, \up{z}]$ in the same way, keeping $\sigma(a) \le \up{\sigma}' < \sigma(b)$. Then $\up{z}' = b$.
\end{enumerate}
Then $[\lo{z}', \up{z}'] \subseteq [\lo{z}, \up{z}]$ contains every $z \in [\lo{z}, \up{z}]$ with $\sigma(z) \in [\lo{\sigma}', \up{\sigma}']$. An empty output means that no such $z$ exists. Each halving divides the width of the bracket $[a, b]$, which bounds the distance of a moved end to the exact crossing, by two. For a certified non-increasing neuron, apply the lemma to $-\sigma$ and $[-\up{\sigma}', -\lo{\sigma}']$.
\end{lemma}

\subsection{Soundness of Backward Contraction}

Algorithm~\ref{alg:tighten} synthesizes these backward propagation and inversion procedures into a unified process. Specifically, line 7 leverages the lower linear bound derived in Theorem~\ref{thm:linear}, which is mathematically validated across the entire spatial region $\Rcal(P)$.

\begin{algorithm}[t]
\caption{Backward contraction of a pattern}
\label{alg:tighten}
\begin{algorithmic}[1]
\Require Pattern $P$; its bounds $\Bcal_P$ with intervals $[\lo{\z}^{(r)}, \up{\z}^{(r)}]$, $[\lo{\sigma}^{(r)}, \up{\sigma}^{(r)}]$ and monotonicity certificates; the linear lower bound $A \x + \cbf_{\mathrm{lo}}$ of Theorem~\ref{thm:linear}
\Ensure Contracted pattern $P'$, or \textsc{empty}
\State Contract $[\lo{\sigma}^{(R)}, \up{\sigma}^{(R)}]$ to $[\lo{\sigma}'^{(R)}, \up{\sigma}'^{(R)}]$ with $H W_{R+1} \sigma^{(R)}(\z^{(R)}) + H \bbf_{R+1} \le h$ \Comment{Lemma~\ref{lem:rows}}
\For{$r = R, \dots, 1$}
    \State For every certified monotone neuron: $[\lo{z}'^{(r)}_j, \up{z}'^{(r)}_j] \gets$ the contracted inverse interval for $[\lo{\sigma}'^{(r)}_j, \up{\sigma}'^{(r)}_j]$; other neurons keep $[\lo{z}^{(r)}_j, \up{z}^{(r)}_j]$ \Comment{Lemma~\ref{lem:inverse}}
    \State Contract $[\lo{\sigma}^{(r-1)}, \up{\sigma}^{(r-1)}]$ to $[\lo{\sigma}'^{(r-1)}, \up{\sigma}'^{(r-1)}]$ with $\lo{\z}'^{(r)} \le W_r \sigma^{(r-1)}(\z^{(r-1)}) + \bbf_r \le \up{\z}'^{(r)}$ \Comment{Lemma~\ref{lem:rows}; $\sigma^{(0)}(\z^{(0)}) = \x$}
    \State \textbf{if} some interval is empty \Return \textsc{empty}
\EndFor
\State Contract $[\lo{\sigma}'^{(0)}, \up{\sigma}'^{(0)}]$ with $A \x + \cbf_{\mathrm{lo}} \le h$; \textbf{if} empty \Return \textsc{empty} \Comment{Lemma~\ref{lem:rows}, Theorem~\ref{thm:linear}}
\State \Return $P'$ with box $\hat X' = [\lo{\sigma}'^{(0)}, \up{\sigma}'^{(0)}]$ and intervals $\hat Z'^{(r)} = \hat Z^{(r)} \cap [\lo{\z}'^{(r)}, \up{\z}'^{(r)}]$
\end{algorithmic}
\end{algorithm}

\begin{theorem}[Soundness of contraction]
\label{thm:tighten}
Let $P'$ be returned by Algorithm~\ref{alg:tighten}. Then $\Rcal(P) \cap \BRS_f(\Y) \subseteq \Rcal(P') \subseteq \Rcal(P)$. If the algorithm returns \textsc{empty}, then $\Rcal(P) \cap \BRS_f(\Y) = \emptyset$.
\end{theorem}

In other words, the iterative contraction sequence delineated in Algorithm~\ref{alg:pattern} preserves mathematical soundness as shown in Theorem~\ref{thm:tighten}, ensuring the recursively composed mapping remains strictly certified. The bounding limits $L$, $U$, and $[\underline{\rho},\overline{\rho}]$ calculated during the terminal bounding pass inherently apply to the established geometric region of that specific pass, and by necessary extension, safely enclose the final contracted subspace. These established limits systematically dictate the final analytical classification of the pattern.

\subsection{Geometric Approximations and Polytope Verification}
The last step is to classify the activation pattern $P$, based on the contracted pattern $P'$, into those that need more processing (undecided regions), those who are part of the inner BRS, and those that are part of the outer BRS. The following result provide the conditions needed for such classification.

\begin{theorem}[Classification of Activation Patterns]
\label{thm:polytopes}
Let $\Bcal_P$, $L$, $U$, $[\lo\rho, \up\rho]$ be the bounds of forward bounding for $P$.
\begin{enumerate}
    \item \emph{Prune.} If $\Bcal_P$ is empty, or $\lo\rho_i > h_i$ for some row $i$, then $\Rcal(P) \cap \BRS_f(\Y) = \emptyset$.
    \item \emph{Accept.} If $\up\rho_i \le h_i$ for every row $i$, then $\Rcal(P) \subseteq \BRS_f(\Y)$.
    \item \emph{Inner part.} $I(P) = \Rcal(P) \cap \{\x : U(\x) \le h\} \subseteq \BRS_f(\Y)$.
    \item \emph{Outer part.} $O(P) = \Rcal(P) \cap \{\x : L(\x) \le h\} \supseteq \Rcal(P) \cap \BRS_f(\Y)$, and $O(P)$ lies in the polytope:
    \begin{equation*}
        \hat O(P) = \big\{\x \in \hat X : \lo{\hat\z}^{(1)} \le W_1 \x + \bbf_1 \le \up{\hat\z}^{(1)}, \; A \x + \cbf_{\mathrm{lo}} \le h \big\}.
    \end{equation*}
\end{enumerate}
\end{theorem}

Every finalized subset allocated to collections $I$ and $O$ adheres to the rigorous structural prerequisites of Problem 1, formulated as a pattern region intersected with $m_{H}$ specific halfspaces. The outer approximations can be efficiently mapped and stored as closed polytopes $\hat{O}(P)$. Upon certifying a pattern as fully accepted, the \tool framework executes an accelerated forward bounding pass that explicitly omits the interval constraints for layers $r\ge2$. If this unconstrained analytical pass verifies that all target rows remain definitively satisfied, the algorithm formalizes the accepted region as an encompassing polytope. This final structural polytope completely encapsulates the original sub-domain $\mathcal{R}(P)$ and, according to item 2 evaluated over the relaxed pattern, is strictly proven to reside within $BRS_{f}(\mathcal{Y})$.

\section{Branching}
\label{sec:refinement}

This section gives the rule of branching for an undecided contracted pattern $P'$ with box $\hat X'$ and an outer polytope $\hat O(P')$ (Theorem~\ref{thm:polytopes}). Appendix~\ref{app:proofs:phase3} gives the proofs and the supporting statements.

\subsection{Candidate and Gap Decomposition}
\label{sec:candidates}\label{sec:gapdecomp}

The first step of splitting is to find the target row with the largest gap between its bounds. To decide which rows are ambiguous, we start with the centre point of $\hat X'$. It is taken and projected in up to $30$ projection steps onto the violated halfspaces of $\hat O(P')$. Then it is clipped to the box after each step. The result is the candidate point $\x^\star$ that is used to select the rows. A row is called ambiguous at $\x^\star$ if $L_i(\x^\star) \le h_i < U_i(\x^\star)$. The deciding row $i^\star$ is the ambiguous row with the largest gap $\delta_i = U_i - L_i$. If no row is ambiguous, $i^\star$ is the row with the largest gap among the rows whose upper row bound $\up\rho_i$ exceeds $h_i$.

After choosing $i^\star$, \tool records the contribution of each neuron to its gap. By the gap identity~\eqref{eq:gap_identity}, the contribution of neuron $(r,j)$ is $\theta_{rj} = |\Lambda^{(r)}_{i^\star j}|\, \beta^{(r)}_j$, the sensitivity of the target row gap to the neuron times the relaxation width of the neuron. These contributions sum to the target gap $\delta_{i^\star}$.

\subsection{Split Selection}
\label{sec:split}

Halving the interval of neuron $(r,j)$ divides its relaxation width $\beta^{(r)}_j$ by about four (Proposition~\ref{prop:band}), so the reduction of the neuron's share of the target gap is about $\tfrac34\theta_{rj}$, which is the neuron score. Halving input dimension $i$ of $\hat X'$ reduces the target gap through every neuron. It shrinks the interval width of $z^{(r)}_j$ by about the factor $1 - S^{(r)}_{ji}/2$, where $S^{(r)}_{ji}$ is the share of that width due to $x_i$ (Appendix~\ref{app:s:shares}). The relaxation width then shrinks by about the factor $(1 - S^{(r)}_{ji}/2)^2$, so the neuron's share of the target gap falls by about $\theta_{rj} \big(1 - (1 - S^{(r)}_{ji}/2)^2\big)$. The sum of these reductions over all neurons is the input score of dimension $i$. \tool compares the largest neuron score with the largest input score and splits the better option (Algorithm~\ref{alg:refine}). For each neuron, the shares of all input dimensions sum to one, so each share is of order $1/d$ for dense weights. An input split therefore tends to win in low dimension, and a neuron split in high dimension. No parameter of the rule depends on $d$.

\begin{algorithm}[t]
\caption{$\textsc{Refine}(P', L, U)$ (branching)}
\label{alg:refine}
\begin{algorithmic}[1]
\Require Undecided contracted pattern $P'$, linear bounds with coefficients $\Lambda^{(r)}$ and relaxation widths $\beta^{(r)}$
\Ensure Children whose regions cover $\Rcal(P')$, with estimated volumes, or $\emptyset$
\State $\x^\star \gets$ projection steps from the box centre onto $\hat O(P')$
\State $i^\star \gets$ deciding row at $\x^\star$; \; $\theta_{rj} \gets |\Lambda^{(r)}_{i^\star j}| \beta^{(r)}_j$ for all neurons~\eqref{eq:gap_identity}
\State $(r, j) \gets \arg\max \tfrac34 \theta_{rj}$; \; $i \gets \arg\max_i \sum_{r,j} \theta_{rj}(1 - (1 - S^{(r)}_{ji}/2)^2)$
\State Draw $M$ uniform samples of $\hat X'$ and keep the $n_{\mathrm{in}}$ samples that lie in $\Rcal(P')$ \Comment{Section~\ref{sec:setdecomp}}
\If{the input score of $i$ exceeds the neuron score of $(r,j)$}
    \State $C \gets$ the two halves of $\hat X'$ along $i$
\ElsIf{the set $S$ of Section~\ref{sec:setdecomp} is non-empty}
    \State $C \gets$ the at most $2|S|+1$ children of the joint split~\eqref{eq:setdec}
\Else
    \State $C \gets$ the split of $\hat Z^{(r)}_j$ at the critical point closest to its midpoint, else at the midpoint
\EndIf
\State Attach to each child its estimated volume~\eqref{eq:mass}; \Return $C$
\end{algorithmic}
\end{algorithm}

\begin{figure*}[t]
\centering
\includegraphics[width=0.95\textwidth]{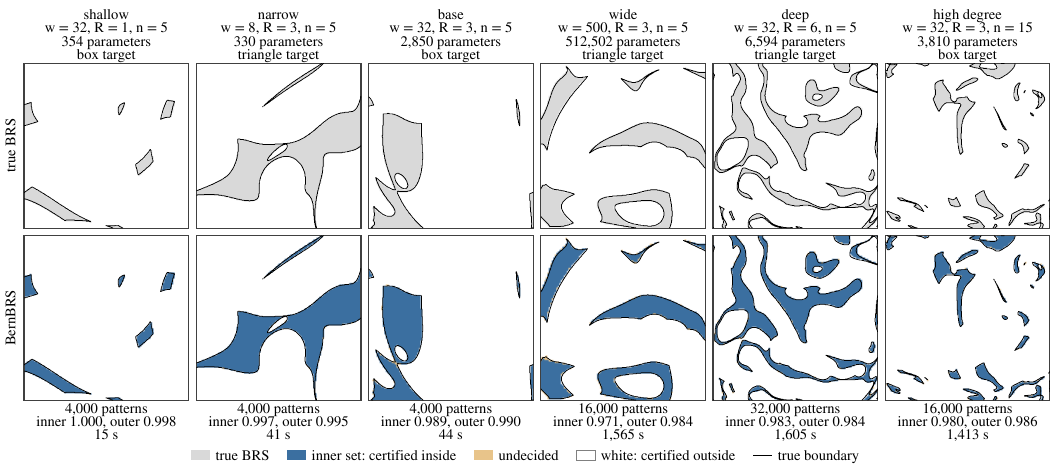}
\caption{True BRS and sets computed by \tool on six random two-input networks. The top row shows the true BRS. The bottom row shows the inner set, the undecided set and the true boundary on the same grid of $400 \times 400$ points. Under each lower panel are the number of activation patterns, the inner and outer ratios, and the seconds of that run on one CPU core. 
}
\vspace{-5mm}
\label{fig:2d:main}
\end{figure*}

If the neuron wins and the set $S$ of Section~\ref{sec:setdecomp} is empty, Algorithm~\ref{alg:refine} splits the best neuron at the candidate closest to the midpoint of $(\lo{z}, \up{z})$, or at the midpoint if there is none. The candidates are the real roots of $(\sigma^{(r)}_j)'$ in the interval if the neuron is not certified monotone. On a fine enough subdivision at these points, the monotonicity certificate holds on every piece (Proposition~\ref{prop:monotone}). Intervals narrower than $10^{-9}$ are never split, and the depth is limited to $D = 100{,}000$.

\subsection{Joint Splitting}
\label{sec:setdecomp}

The joint split rests on the concentration of the pre-activations. A pre-activation interval is as wide as the sum of the weight magnitudes, while the pre-activations of almost all inputs lie within a few standard deviations of their mean, and the standard deviation is smaller than the width by a factor between $\sqrt{12}$ and $\sqrt{12 d}$ (Proposition~\ref{prop:concentration}). A bisection of such an interval halves it only and leaves almost all of the volume in one child. Restricting the neuron to the narrow part shrinks its interval far more and moves almost no volume into the two tails.

The narrow part is found from samples. Algorithm~\ref{alg:refine} draws $M$ uniform samples of $\hat X'$ and keeps those that lie in $\Rcal(P')$, at least $16$ of them. For a neuron, the mean $\mu$ and the standard deviation $\varsigma$ of its pre-activation over these samples define its \emph{central interval} $[\mu - \kappa\varsigma, \mu + \kappa\varsigma] \cap [\lo{z}, \up{z}]$. The neuron is \emph{eligible} if this interval is narrower than $\omega(\up{z} - \lo{z})$. We use $\kappa = 4$ and $\omega = 1/2$.

The neurons to restrict come from the ranking of Section~\ref{sec:split}. Sorted by score $\tfrac34\theta_{rj}$, decreasing, the shortest prefix whose scores sum to at least $\varphi = 0.8$ times the total is selected, and its eligible neurons $(r_1, j_1), \dots, (r_{|S|}, j_{|S|})$ form the set $S$, in that order.

Let $C_t = [\underline c_t, \overline c_t]$ be the central interval and $z_t$ the pre-activation of the $t$-th neuron. The joint split nests the restrictions. The first neuron gives two tail children. With the first neuron held in its central interval, the second neuron gives two more, and so on. The last child holds every neuron of $S$ in its central interval. The children of $P'$ are
\begin{equation}
\begin{aligned}
T_t^{-} &= \Rcal(P') \cap \{ z_s \in C_s,\ s < t \} \cap \{ z_t \le \underline c_t \}, \\
T_t^{+} &= \Rcal(P') \cap \{ z_s \in C_s,\ s < t \} \cap \{ z_t \ge \overline c_t \}, \\
K &= \Rcal(P') \cap \{ z_s \in C_s,\ s = 1, \dots, |S| \}.
\end{aligned}
\label{eq:setdec}
\end{equation}
Here $T_t^{\pm}$, $t = 1, \dots, |S|$, are the tails of the $t$-th neuron and $K$ is the central child. These at most $2|S| + 1$ children cover $\Rcal(P')$ (Lemma~\ref{lem:setdecomp}). An input of $\Rcal(P')$ either has every $z_t$ in $C_t$ and lies in $K$, or it has a first neuron $t$ with $z_t$ outside $C_t$ and lies in $T_t^-$ or $T_t^+$. Each child carries an estimated volume, which only orders the priority queue. A tail receives about $\eta = \Phi(-\kappa) \approx 3 \cdot 10^{-5}$ of its parent (Appendix~\ref{app:s:mass}). The joint split spends one processed pattern and one round on the whole set.

\section{Experiments}
\label{sec:experiments}

We evaluate the proposed \tool framework through two foundational experiments and two applied case studies. The first experiment assesses the algorithmic correctness of the framework on randomized two-input networks. The second experiment investigates the framework's scalability when applied to image classifiers containing up to 2.3 million parameters. The first case study examines a hybrid dynamical system, specifically a legged robot operating in a closed loop. The second case study evaluates a distilled transformer block based on the TinyBERT architecture. Performance is quantified using the \emph{inner ratio} $\vol(\bigcup \Ical) / \vol(\BRS_f(\Y))$ and the \emph{outer ratio} $\vol(\BRS_f(\Y)) / \vol(\bigcup \Ocal)$, which evaluate the volumes of the estimated sets relative to the true backward reachable set (BRS), yielding a value of 1 for exact approximations. Volumes are approximated using uniform samples from the input space.

\begin{table*}[t]
\centering
\caption{
Performance comparison of \tool versus the CROWN baseline with a time-matched budget. The data details inner ratios, outer ratios, solving times, and model parameters evaluated over three test images.
}
\label{tab:images:scale}
\scriptsize\setlength{\tabcolsep}{2.5pt}\begin{tabular}{llrrrrrrrrrrrrr}
\toprule
 & & & & & \multicolumn{5}{c}{\tool} & \multicolumn{5}{c}{CROWN} \\
\cmidrule(lr){6-10}\cmidrule(lr){11-15}
Data set & \shortstack{Bernstein\\network} & Size & Radius & \# param & IR & OR & T (s) & Patterns & Splits & IR & OR & T (s) & Elements & Splits \\
\midrule
MNIST & 1 & $100 \times 4$ & $8/255$ & 113,810 & {\boldmath\bfseries 0.634 $\pm$ 0.076} & {\boldmath\bfseries 0.715 $\pm$ 0.038} & 16.3 $\pm$ 0.2 & 1,259 $\pm$ 47 & 400 $\pm$ 0 & 0.000 $\pm$ 0.000 & 0.501 $\pm$ 0.004 & 13.8 $\pm$ 0.0 & 8,192 $\pm$ 0 & 8,191 $\pm$ 0 \\
 &  & $300 \times 4$ & $8/255$ & 521,410 & {\boldmath\bfseries 0.543 $\pm$ 0.028} & {\boldmath\bfseries 0.656 $\pm$ 0.035} & 34.2 $\pm$ 1.0 & 2,809 $\pm$ 237 & 400 $\pm$ 0 & 0.000 $\pm$ 0.000 & 0.500 $\pm$ 0.007 & 31.5 $\pm$ 0.0 & 8,192 $\pm$ 0 & 8,191 $\pm$ 0 \\
 &  & $500 \times 4$ & $8/255$ & 1,169,010 & {\boldmath\bfseries 0.482 $\pm$ 0.086} & {\boldmath\bfseries 0.625 $\pm$ 0.021} & 72.6 $\pm$ 3.4 & 4,278 $\pm$ 273 & 400 $\pm$ 0 & 0.000 $\pm$ 0.000 & 0.501 $\pm$ 0.006 & 55.6 $\pm$ 0.3 & 8,192 $\pm$ 0 & 8,191 $\pm$ 0 \\
\addlinespace[1pt]
 &  & $100 \times 2$ & 0.02 & 91,610 & {\boldmath\bfseries 0.845 $\pm$ 0.011} & {\boldmath\bfseries 0.851 $\pm$ 0.019} & 6.2 $\pm$ 0.5 & 803 $\pm$ 25 & 400 $\pm$ 0 & 0.000 $\pm$ 0.000 & 0.499 $\pm$ 0.005 & 5.2 $\pm$ 1.1 & 3,413 $\pm$ 1,182 & 3,412 $\pm$ 1,182 \\
 &  & $100 \times 3$ & 0.02 & 102,710 & {\boldmath\bfseries 0.794 $\pm$ 0.009} & {\boldmath\bfseries 0.835 $\pm$ 0.034} & 9.8 $\pm$ 0.8 & 1,020 $\pm$ 32 & 400 $\pm$ 0 & 0.000 $\pm$ 0.000 & 0.498 $\pm$ 0.003 & 7.3 $\pm$ 0.0 & 4,096 $\pm$ 0 & 4,095 $\pm$ 0 \\
 &  & $100 \times 4$ & 0.02 & 113,810 & {\boldmath\bfseries 0.743 $\pm$ 0.045} & {\boldmath\bfseries 0.790 $\pm$ 0.032} & 13.3 $\pm$ 0.3 & 1,253 $\pm$ 34 & 400 $\pm$ 0 & 0.000 $\pm$ 0.000 & 0.501 $\pm$ 0.003 & 8.7 $\pm$ 0.1 & 4,096 $\pm$ 0 & 4,095 $\pm$ 0 \\
\cmidrule(lr){2-15}
 & 2 & $100 \times 4$ & $8/255$ & 113,810 & {\boldmath\bfseries 0.700 $\pm$ 0.054} & {\boldmath\bfseries 0.829 $\pm$ 0.045} & 18.5 $\pm$ 1.6 & 1,039 $\pm$ 193 & 400 $\pm$ 0 & 0.000 $\pm$ 0.000 & 0.497 $\pm$ 0.011 & 13.9 $\pm$ 0.1 & 8,192 $\pm$ 0 & 8,191 $\pm$ 0 \\
 &  & $300 \times 4$ & $8/255$ & 521,410 & {\boldmath\bfseries 0.351 $\pm$ 0.091} & {\boldmath\bfseries 0.624 $\pm$ 0.038} & 33.2 $\pm$ 2.4 & 2,040 $\pm$ 349 & 400 $\pm$ 0 & 0.000 $\pm$ 0.000 & 0.497 $\pm$ 0.004 & 28.2 $\pm$ 6.3 & 6,827 $\pm$ 2,365 & 6,826 $\pm$ 2,365 \\
 &  & $500 \times 4$ & $8/255$ & 1,169,010 & {\boldmath\bfseries 0.141 $\pm$ 0.147} & {\boldmath\bfseries 0.548 $\pm$ 0.063} & 62.6 $\pm$ 2.6 & 2,625 $\pm$ 309 & 400 $\pm$ 0 & 0.000 $\pm$ 0.000 & 0.498 $\pm$ 0.002 & 55.4 $\pm$ 0.3 & 8,192 $\pm$ 0 & 8,191 $\pm$ 0 \\
\addlinespace[1pt]
 &  & $100 \times 2$ & 0.02 & 91,610 & {\boldmath\bfseries 0.838 $\pm$ 0.020} & {\boldmath\bfseries 0.904 $\pm$ 0.028} & 5.8 $\pm$ 0.6 & 720 $\pm$ 22 & 400 $\pm$ 0 & 0.000 $\pm$ 0.000 & 0.506 $\pm$ 0.020 & 4.5 $\pm$ 1.2 & 2,731 $\pm$ 1,182 & 2,730 $\pm$ 1,182 \\
 &  & $100 \times 3$ & 0.02 & 102,710 & {\boldmath\bfseries 0.773 $\pm$ 0.069} & {\boldmath\bfseries 0.855 $\pm$ 0.053} & 10.2 $\pm$ 1.3 & 759 $\pm$ 143 & 398 $\pm$ 3 & 0.000 $\pm$ 0.000 & 0.498 $\pm$ 0.010 & 7.4 $\pm$ 0.3 & 4,096 $\pm$ 0 & 4,095 $\pm$ 0 \\
 &  & $100 \times 4$ & 0.02 & 113,810 & {\boldmath\bfseries 0.798 $\pm$ 0.053} & {\boldmath\bfseries 0.881 $\pm$ 0.034} & 17.6 $\pm$ 1.9 & 1,066 $\pm$ 235 & 400 $\pm$ 0 & 0.000 $\pm$ 0.000 & 0.497 $\pm$ 0.010 & 13.8 $\pm$ 0.1 & 8,192 $\pm$ 0 & 8,191 $\pm$ 0 \\
\midrule
GTSRB & 1 & $100 \times 4$ & $8/255$ & 345,943 & {\boldmath\bfseries 0.429 $\pm$ 0.033} & {\boldmath\bfseries 0.672 $\pm$ 0.007} & 16.4 $\pm$ 0.2 & 1,283 $\pm$ 48 & 400 $\pm$ 0 & 0.000 $\pm$ 0.000 & 0.489 $\pm$ 0.005 & 13.0 $\pm$ 0.1 & 2,048 $\pm$ 0 & 2,047 $\pm$ 0 \\
 &  & $300 \times 4$ & $8/255$ & 1,217,743 & {\boldmath\bfseries 0.448 $\pm$ 0.038} & {\boldmath\bfseries 0.658 $\pm$ 0.012} & 69.3 $\pm$ 0.6 & 2,917 $\pm$ 11 & 400 $\pm$ 0 & 0.000 $\pm$ 0.000 & 0.499 $\pm$ 0.003 & 68.2 $\pm$ 0.3 & 8,192 $\pm$ 0 & 8,191 $\pm$ 0 \\
 &  & $500 \times 4$ & $8/255$ & 2,329,543 & {\boldmath\bfseries 0.372 $\pm$ 0.056} & {\boldmath\bfseries 0.642 $\pm$ 0.011} & 137.3 $\pm$ 1.1 & 4,436 $\pm$ 43 & 400 $\pm$ 0 & 0.000 $\pm$ 0.000 & 0.499 $\pm$ 0.003 & 116.3 $\pm$ 5.5 & 6,144 $\pm$ 3,547 & 6,143 $\pm$ 3,547 \\
\addlinespace[1pt]
 &  & $100 \times 2$ & 0.02 & 323,743 & {\boldmath\bfseries 0.731 $\pm$ 0.029} & {\boldmath\bfseries 0.839 $\pm$ 0.007} & 8.4 $\pm$ 0.1 & 816 $\pm$ 8 & 400 $\pm$ 0 & 0.000 $\pm$ 0.000 & 0.490 $\pm$ 0.003 & 6.9 $\pm$ 0.1 & 1,024 $\pm$ 0 & 1,023 $\pm$ 0 \\
 &  & $100 \times 3$ & 0.02 & 334,843 & {\boldmath\bfseries 0.659 $\pm$ 0.047} & {\boldmath\bfseries 0.789 $\pm$ 0.026} & 14.0 $\pm$ 0.4 & 1,064 $\pm$ 72 & 400 $\pm$ 0 & 0.000 $\pm$ 0.000 & 0.492 $\pm$ 0.007 & 12.6 $\pm$ 0.3 & 2,048 $\pm$ 0 & 2,047 $\pm$ 0 \\
 &  & $100 \times 4$ & 0.02 & 345,943 & {\boldmath\bfseries 0.607 $\pm$ 0.023} & {\boldmath\bfseries 0.754 $\pm$ 0.008} & 16.2 $\pm$ 0.1 & 1,244 $\pm$ 13 & 400 $\pm$ 0 & 0.000 $\pm$ 0.000 & 0.489 $\pm$ 0.006 & 13.9 $\pm$ 0.1 & 2,048 $\pm$ 0 & 2,047 $\pm$ 0 \\
\cmidrule(lr){2-15}
 & 2 & $100 \times 4$ & $8/255$ & 345,943 & {\boldmath\bfseries 0.161 $\pm$ 0.057} & {\boldmath\bfseries 0.575 $\pm$ 0.043} & 19.5 $\pm$ 2.4 & 1,978 $\pm$ 583 & 400 $\pm$ 0 & 0.000 $\pm$ 0.000 & 0.499 $\pm$ 0.004 & 14.2 $\pm$ 0.5 & 2,048 $\pm$ 0 & 2,047 $\pm$ 0 \\
 &  & $300 \times 4$ & $8/255$ & 1,217,743 & {\boldmath\bfseries 0.046 $\pm$ 0.015} & {\boldmath\bfseries 0.513 $\pm$ 0.011} & 63.1 $\pm$ 2.5 & 2,606 $\pm$ 214 & 400 $\pm$ 0 & 0.000 $\pm$ 0.000 & 0.505 $\pm$ 0.007 & 41.3 $\pm$ 0.2 & 4,096 $\pm$ 0 & 4,095 $\pm$ 0 \\
 &  & $500 \times 4$ & $8/255$ & 2,329,543 & {\boldmath\bfseries 0.060 $\pm$ 0.058} & {\boldmath\bfseries 0.538 $\pm$ 0.008} & 126.4 $\pm$ 42.9 & 4,043 $\pm$ 1,809 & 400 $\pm$ 0 & 0.000 $\pm$ 0.000 & 0.505 $\pm$ 0.010 & 112.7 $\pm$ 23.1 & 1,579 $\pm$ 2,189 & 1,578 $\pm$ 2,189 \\
\addlinespace[1pt]
 &  & $100 \times 2$ & 0.02 & 323,743 & {\boldmath\bfseries 0.454 $\pm$ 0.184} & {\boldmath\bfseries 0.677 $\pm$ 0.048} & 8.4 $\pm$ 0.3 & 790 $\pm$ 32 & 400 $\pm$ 0 & 0.000 $\pm$ 0.000 & 0.486 $\pm$ 0.006 & 7.4 $\pm$ 0.1 & 1,024 $\pm$ 0 & 1,023 $\pm$ 0 \\
 &  & $100 \times 3$ & 0.02 & 334,843 & {\boldmath\bfseries 0.416 $\pm$ 0.093} & {\boldmath\bfseries 0.648 $\pm$ 0.037} & 12.5 $\pm$ 0.8 & 1,073 $\pm$ 88 & 400 $\pm$ 0 & 0.000 $\pm$ 0.000 & 0.490 $\pm$ 0.010 & 10.3 $\pm$ 2.5 & 1,365 $\pm$ 591 & 1,364 $\pm$ 591 \\
 &  & $100 \times 4$ & 0.02 & 345,943 & {\boldmath\bfseries 0.384 $\pm$ 0.068} & {\boldmath\bfseries 0.652 $\pm$ 0.048} & 20.8 $\pm$ 3.2 & 2,045 $\pm$ 819 & 400 $\pm$ 0 & 0.000 $\pm$ 0.000 & 0.499 $\pm$ 0.004 & 14.2 $\pm$ 0.5 & 2,048 $\pm$ 0 & 2,047 $\pm$ 0 \\
\midrule
CIFAR-10 & 1 & $100 \times 4$ & $8/255$ & 342,610 & {\boldmath\bfseries 0.649 $\pm$ 0.069} & {\boldmath\bfseries 0.741 $\pm$ 0.048} & 14.9 $\pm$ 0.5 & 1,267 $\pm$ 107 & 400 $\pm$ 0 & 0.000 $\pm$ 0.000 & 0.504 $\pm$ 0.008 & 13.8 $\pm$ 0.4 & 4,096 $\pm$ 0 & 4,095 $\pm$ 0 \\
 &  & $300 \times 4$ & $8/255$ & 1,207,810 & {\boldmath\bfseries 0.515 $\pm$ 0.044} & {\boldmath\bfseries 0.710 $\pm$ 0.050} & 69.0 $\pm$ 1.5 & 2,896 $\pm$ 101 & 400 $\pm$ 0 & 0.000 $\pm$ 0.000 & 0.502 $\pm$ 0.005 & 49.9 $\pm$ 0.3 & 8,192 $\pm$ 0 & 8,191 $\pm$ 0 \\
 &  & $500 \times 4$ & $8/255$ & 2,313,010 & {\boldmath\bfseries 0.476 $\pm$ 0.069} & {\boldmath\bfseries 0.681 $\pm$ 0.054} & 137.4 $\pm$ 3.3 & 4,397 $\pm$ 105 & 400 $\pm$ 0 & 0.000 $\pm$ 0.000 & 0.500 $\pm$ 0.016 & 112.0 $\pm$ 16.1 & 8,192 $\pm$ 0 & 8,191 $\pm$ 0 \\
\addlinespace[1pt]
 &  & $100 \times 2$ & 0.02 & 320,410 & {\boldmath\bfseries 0.800 $\pm$ 0.043} & {\boldmath\bfseries 0.852 $\pm$ 0.042} & 7.3 $\pm$ 0.2 & 806 $\pm$ 12 & 400 $\pm$ 0 & 0.000 $\pm$ 0.000 & 0.507 $\pm$ 0.006 & 6.2 $\pm$ 0.1 & 2,048 $\pm$ 0 & 2,047 $\pm$ 0 \\
 &  & $100 \times 3$ & 0.02 & 331,510 & {\boldmath\bfseries 0.765 $\pm$ 0.034} & {\boldmath\bfseries 0.832 $\pm$ 0.027} & 11.2 $\pm$ 0.2 & 1,075 $\pm$ 51 & 400 $\pm$ 0 & 0.000 $\pm$ 0.000 & 0.504 $\pm$ 0.010 & 8.1 $\pm$ 0.6 & 2,048 $\pm$ 0 & 2,047 $\pm$ 0 \\
 &  & $100 \times 4$ & 0.02 & 342,610 & {\boldmath\bfseries 0.766 $\pm$ 0.043} & {\boldmath\bfseries 0.812 $\pm$ 0.035} & 15.1 $\pm$ 0.5 & 1,241 $\pm$ 58 & 400 $\pm$ 0 & 0.000 $\pm$ 0.000 & 0.504 $\pm$ 0.007 & 14.0 $\pm$ 0.4 & 4,096 $\pm$ 0 & 4,095 $\pm$ 0 \\
\cmidrule(lr){2-15}
 & 2 & $100 \times 4$ & $8/255$ & 342,610 & {\boldmath\bfseries 0.246 $\pm$ 0.111} & {\boldmath\bfseries 0.572 $\pm$ 0.053} & 14.4 $\pm$ 1.4 & 746 $\pm$ 17 & 400 $\pm$ 0 & 0.000 $\pm$ 0.000 & 0.495 $\pm$ 0.006 & 11.8 $\pm$ 2.0 & 3,413 $\pm$ 1,182 & 3,412 $\pm$ 1,182 \\
 &  & $300 \times 4$ & $8/255$ & 1,207,810 & {\boldmath\bfseries 0.035 $\pm$ 0.029} & {\boldmath\bfseries 0.516 $\pm$ 0.005} & 49.8 $\pm$ 2.6 & 1,555 $\pm$ 68 & 400 $\pm$ 0 & 0.000 $\pm$ 0.000 & 0.498 $\pm$ 0.007 & 37.1 $\pm$ 11.2 & 5,461 $\pm$ 2,365 & 5,460 $\pm$ 2,365 \\
 &  & $500 \times 4$ & $8/255$ & 2,313,010 & {\boldmath\bfseries 0.018 $\pm$ 0.010} & {\boldmath\bfseries 0.508 $\pm$ 0.013} & 88.0 $\pm$ 1.6 & 2,332 $\pm$ 72 & 400 $\pm$ 0 & 0.000 $\pm$ 0.000 & 0.497 $\pm$ 0.006 & 72.5 $\pm$ 10.8 & 3,413 $\pm$ 1,182 & 3,412 $\pm$ 1,182 \\
\addlinespace[1pt]
 &  & $100 \times 2$ & 0.02 & 320,410 & {\boldmath\bfseries 0.430 $\pm$ 0.047} & {\boldmath\bfseries 0.655 $\pm$ 0.030} & 7.4 $\pm$ 0.4 & 815 $\pm$ 55 & 400 $\pm$ 0 & 0.000 $\pm$ 0.000 & 0.497 $\pm$ 0.010 & 5.9 $\pm$ 0.0 & 2,048 $\pm$ 0 & 2,047 $\pm$ 0 \\
 &  & $100 \times 3$ & 0.02 & 331,510 & {\boldmath\bfseries 0.511 $\pm$ 0.016} & {\boldmath\bfseries 0.681 $\pm$ 0.038} & 10.1 $\pm$ 0.1 & 762 $\pm$ 2 & 400 $\pm$ 0 & 0.000 $\pm$ 0.000 & 0.500 $\pm$ 0.007 & 7.2 $\pm$ 0.0 & 2,048 $\pm$ 0 & 2,047 $\pm$ 0 \\
 &  & $100 \times 4$ & 0.02 & 342,610 & {\boldmath\bfseries 0.452 $\pm$ 0.117} & {\boldmath\bfseries 0.647 $\pm$ 0.059} & 14.5 $\pm$ 1.1 & 746 $\pm$ 28 & 400 $\pm$ 0 & 0.000 $\pm$ 0.000 & 0.496 $\pm$ 0.006 & 13.0 $\pm$ 0.2 & 4,096 $\pm$ 0 & 4,095 $\pm$ 0 \\
\bottomrule
\end{tabular}

\vspace{-5mm}
\end{table*}

~\\\noindent$\bullet$ \textbf{Experiment 1: Correctness.}
This experiment explicitly validates the mathematical correctness and geometric accuracy of the \tool framework. We utilize random, untrained two-input neural networks with two outputs and randomly assigned Bernstein activations. The true BRS is computed over a fine grid to serve as an exact ground truth for comparison against the algorithm's computed bounds. Targets are defined as geometric boxes and triangles.

Figure~\ref{fig:2d:main} shows six networks of different width $w$, depth $R$ and degree $n$.  As shown in the figure, \tool demonstrated strict soundness across 859 independent trials, maintaining inner ratios of at least 0.97 and outer ratios of at least 0.98 for select network configurations. Crucially, no elements of the generated inner sets fell outside the true BRS, and no elements of the true BRS fell outside the outer sets. While median inner ratios naturally decreased for deeper or wider architectures under strict computational budgets, scaling the allowed partitions to 32,000 consistently restored inner and outer ratios to 0.90.

~\\\noindent$\bullet$ \textbf{Experiment 2: Scalability Against State-of-the-Art.}
This experiment measures the computational scalability of the framework across increasingly complex neural network architectures. The baseline method for comparison is the backward reachability technique in~\cite{rober2023} which uses CROWN-based linear relaxations of neural networks~\cite{zhang2018crown} along with input space partitioning. That method computes over-approximations, and we add the corresponding under-approximation.
We implement it for Bernstein layers and compare it with \tool on the image classifiers and the closed loop, on the same networks, samples and hardware. We did not use PREMAP~\cite{premap2025} or INVPROP~\cite{kotha2023provably} as baselines, because both are formulated for ReLU networks and their published versions do not cover Bernstein activations.

The framework is tested on fully connected Bernstein image classifiers trained on MNIST~\cite{lecun1998mnist}, GTSRB~\cite{stallkamp2012gtsrb} and CIFAR-10~\cite{krizhevsky2009} datasets. The input domains are defined as bounding boxes around test images, and targets are formulated to keep the label margin above its median value. \tool is benchmarked directly against the CROWN-based baseline restricted to an equivalent time allocation.
Appendix~\ref{app:exp:images} gives the details of their training. Each run uses an $\ell_\infty$ box around one of three test images, with half-width $8/255$ across widths and $0.02$ across depths. The target keeps the label margin above its median over the box, so the true share is about one half. The baseline gets the time of \tool: its element budget doubles from one, and we keep the last baseline run within the solve time of \tool on that box.

Table~\ref{tab:images:scale} gives the results. The baseline is never ahead. It has an inner ratio of $0$ on all $108$ runs and its outer set is the whole box on $105$, although it makes a median of $4{,}095$ splits against about $400$ for \tool. From width $100$ to $500$, the inner ratio of \tool falls from $0.43$--$0.65$ to $0.37$--$0.48$ on Bernstein network~1 and from $0.16$--$0.70$ to $0.02$--$0.14$ on Bernstein network~2. From depth $2$ to $4$, it falls from $0.73$--$0.84$ to $0.61$--$0.77$ on Bernstein network~1 and from $0.43$--$0.84$ to $0.38$--$0.80$ on Bernstein network~2. The solve takes at most $176$\,s per run. The test accuracy is at least $96.6$\,\% on MNIST, $88.9$\,\% on GTSRB and $48.2$\,\% on CIFAR-10. All runs pass the soundness check.

\begin{table*}[!h]
\caption{
Five text sentences targeting a positive label, evaluated with their free words in bold. The columns contrast the BRS words certified by the \tool student model against words eliciting a positive label from TinyBERT.
}
\label{tab:tinybert:example}
\footnotesize\setlength{\tabcolsep}{3pt}\raggedright
\begin{tabular}{@{}>{\raggedright\arraybackslash}p{0.30\textwidth}>{\raggedright\arraybackslash}p{0.33\textwidth}>{\raggedright\arraybackslash}p{0.33\textwidth}@{}}
\toprule
Sentence & Student & TinyBERT \\
\midrule
i just \textbf{loved} every minute of this film . & proud, enjoyed, loving, amazing & proud, enjoyed, loving, amazing \\
my thoughts were \textbf{focused} on the characters . & concentrate, focus, focusing & concentrate, focus, focusing \\
jones . . . does offer a \textbf{brutal} form of charisma . & ruthless & ruthless \\
a warm but \textbf{realistic} meditation on friendship , family and affection . & accurate, perfection, luxury, positive, progressive, true, uniquely, uncanny, imaginative, solid, beautiful, vision, fine, magnificent, splendid, squared, flawless, exquisite & accurate, perfection, luxury, positive, progressive, true, uniquely, uncanny, imaginative, solid, beautiful, vision, fine, magnificent, splendid, squared, flawless, exquisite \\
one of the most significant moviegoing \textbf{pleasures} of the year . & leisure, enjoy, joy, treat, enjoyable, rewards, happiness, beneficial, delight, belle, charm, pleasure, touch, benefits & leisure, enjoy, joy, treat, enjoyable, rewards, happiness, beneficial, delight, belle, charm, pleasure, touch, benefits \\
\bottomrule
\end{tabular}

\vspace{-3mm}
\end{table*}

~\\\noindent$\bullet$ \textbf{Case Study 1: A Hybrid Dynamical Control System.}
The first case study applies \tool to a trained controller in closed loop. A Bernstein network with one hidden layer of 128 units, trained by PPO~\cite{schulman2017ppo} with certified training, controls HalfCheetah in MuJoCo~\cite{todorov2012mujoco}, a simulated robot with 17 states and 6 actions. We linearize the simulator step along a logged trajectory. For each time step, we merge the one-layer policy with the linearized dynamics, which gives two layers per step, so a $k$-step network has $2k$ layers. We do this up to $k = 20$, which gives a network of depth $40$. Appendix~\ref{app:exp:halfcheetah} gives the details of the merge, the training and the settings.

All backward reachable sets computed in this section are sets of initial states of the linearized closed-loop model for which the merged Bernstein network reaches the specified pitch target. They do not constitute formal safety or failure guarantees for the original MuJoCo plant or for any physical realization of HalfCheetah. Contact changes and model mismatch might lead to discrepancies between the linearized model and MuJoCo at longer horizons; these discrepancies should be understood as unmodeled plant uncertainty that lies outside the scope of the current certification framework which focuses solely on the neural network controller.

After training, we run the controller 50 times for 1,000 steps each, with a deterministic policy and deterministic dynamics, from different random initial states. The robot falls in 8 of these runs, where the pitch passes 1\,rad without recovery, and we analyze 5 of these falls. For each fall, the target is a pitch of at least 1\,rad at the step $E$ two steps after the logged pitch first passes 1\,rad. For $k = 1, \dots, 20$, we compute the BRS over a box around the logged state $k$ steps before $E$, where one step is 0.05\,s. The wide box has half-width $S_{16}$ on all 17 state coordinates, and the thin box keeps $S_{16}$ on the pitch, the pitch rate and the forward speed and uses $S_4$ elsewhere, where $S_m$ is $m$ times a representative sensor error per coordinate (Tables~\ref{tab:halfcheetah:boxes} and~\ref{tab:halfcheetah:scales}). This gives 200 runs at $N = 2{,}000$.

\begin{figure}[t]
\centering
\includegraphics{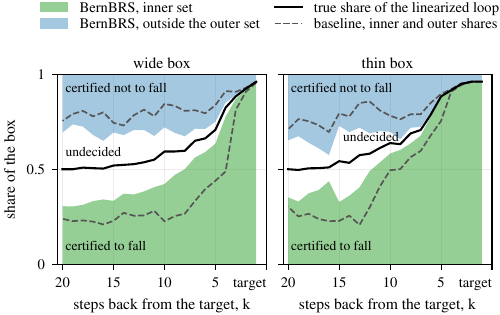}
\caption{
Certificate tightness plotted against the number of backward steps from the target across five failure scenarios. Curves represent box shares: green for the \tool inner set, blue for states outside the outer set, and black for the true share. Dashed lines indicate the CROWN baseline's performance within an equivalent timeframe.
}
\vspace{-5mm}
\label{fig:halfcheetah:main}
\end{figure}

Figure~\ref{fig:halfcheetah:main} shows the shares of the union of the five falls. Green is the inner set, the share of the union that is certified to fall. Blue is the share that lies outside the outer set, which is certified not to fall. The lower dashed line is the inner share of the baseline, and the upper dashed line is its outer share. The black line in the middle is the true share, and the closer the inner and outer shares lie to it, the tighter the BRS is.

We then replay in MuJoCo the states certified to fall, among 4,000 uniform samples of each box, to check them. Figure~\ref{fig:halfcheetah:frames} in the appendix shows a logged fall as an example of such a trajectory.

~\\\noindent$\bullet$ \textbf{Case Study 2: TinyBERT Transformer.}
The second case study applies \tool to a transformer. We take TinyBERT~\cite{jiao2020tinybert} with a classifier, fine-tuned on the sentiment dataset SST-2~\cite{socher2013sst}. We distill its last block and classifier into a Bernstein student, which gives the same prediction as TinyBERT on $99.2\,\%$ of the development sentences. The student has three attention heads with a monotone Bernstein gate of degree $9$ in place of the softmax, an MLP of $300$ Bernstein neurons and $312$ more before the classifier (Appendix~\ref{app:exp:tinybert}).

We take $50$ random development sentences and keep the $43$ that TinyBERT and the student classify correctly. We free one word of each sentence and keep the other words fixed, so the student depends only on the $312$ numbers that represent this word at the input of the block. We then use \tool on the student, through an exact rewrite (Appendix~\ref{app:exp:tinybert:setup}). The input set is a box around the free word, in which each number moves by at most one standard deviation of that number over the training tokens ($\eps\sigma_i$ with $\eps = 1$). We set the target to the true label of the sentence and take the inner set. An input in the inner set is certified to keep that label.

We then test the result on TinyBERT. We take all the candidate words that fall in the $43$ boxes, which are $6{,}321$ words, counted once per box, and place each one in its sentence. \tool proves that $6{,}320$ of these words keep the label on the student, and the remaining word falls in the undecided set. We run TinyBERT on the changed sentences, and all $6{,}321$ words keep the same label. 
This test covers these words only and is not a guarantee for TinyBERT: our backward-reachability certificates apply exclusively to the distilled Bernstein student network obtained by our exact rewrite, under the idealized mathematical model. TinyBERT itself remains unchecked; we only report empirical agreement between TinyBERT and the student on the probed substitutions.
Table~\ref{tab:tinybert:example} shows five sentences with a positive label. In all five, the words proven on the student also give the positive label on TinyBERT, including words with a negative meaning, such as ``ruthless''.

\section*{Acknowledgment}
This work was partially supported by the NSF award CNS \#2504809 and the UCI ProperAI Institute, an Engineering+Society Institute funded as part of a generous gift from Susan and Henry Samueli.

\bibliographystyle{ACM-Reference-Format}
\bibliography{references}

\clearpage
\appendix
\section*{Appendices}

\section{Forward Bounding}
\label{app:proofs:phase1}

This appendix proves the statements of Section~\ref{sec:bounds}. It also states Remark~\ref{rem:subintervals} and Proposition~\ref{prop:band}, which that section cites.

\subsection{Interval Bounds over a Pattern}
\label{app:s:ibp}

\begin{lemma}[Interval bounds over a pattern]
\label{lem:ibp}
Let the bounds be given by~\eqref{eq:ibp_affine} and~\eqref{eq:ibp_act}, with $[\lo{\sigma}^{(0)}, \up{\sigma}^{(0)}] = \hat X$. Then $\z^{(r)}(\x) \in [\lo{\z}^{(r)}, \up{\z}^{(r)}]$ and $\sigma^{(r)}(\z^{(r)}(\x)) \in [\lo{\sigma}^{(r)}, \up{\sigma}^{(r)}]$ for every $\x \in \Rcal(P)$ and $r \in [R]$. If $\lo{z}^{(r)}_j > \up{z}^{(r)}_j$ for some neuron, then $\Rcal(P) = \emptyset$.
\end{lemma}

\begin{proof}[Proof of Lemma~\ref{lem:ibp}]
By induction on $r$. Splitting $W_r$ into its positive and negative parts shows that the first arguments in~\eqref{eq:ibp_affine} enclose $\z^{(r)}(\x)$ for $\x \in \hat X$ whose post-activations lie in the previous bounds. For $\x \in \Rcal(P)$, also $\z^{(r)}(\x) \in \hat Z^{(r)}$ by Definition~\ref{def:pattern}, hence the maximum and minimum. The post-activation bounds follow from Lemma~\ref{lem:clamped}, applied to the interval $[\lo{z}^{(r)}_j, \up{z}^{(r)}_j]$ that contains $z^{(r)}_j(\x)$. If the interval of a neuron is empty, no $\x$ satisfies the constraints.
\end{proof}

\subsection{Bernstein Linear Relaxation}
\label{app:s:relaxation}

\begin{proof}[Proof of Lemma~\ref{lem:relaxation}]
Let $p(z) = \sigma^{(r)}_j(z) - s z$, a polynomial of degree at most $n$. By (P4), the Bernstein coefficients of $z \mapsto s z$ on $[\lo{z}, \up{z}]$ are $s\,\xi_i$. By linearity of the Bernstein form, the Bernstein coefficients of $p$ on this interval are $e_i$. By (P1), $\min_i e_i \le p(z) \le \max_i e_i$ for $z \in [\lo{z}, \up{z}]$.
\end{proof}

For $\up{z} = \lo{z}$ any slope is valid, and the implementation uses the derivative of the activation at $\lo{z}$. With the secant slope, $e_0 = e_n$. Minimizing $\beta$ over the slope $s$ is an option: $\up\beta$ is a maximum and $\lo\beta$ a minimum of functions that are affine in $s$, so $\beta$ is a convex piecewise-linear function of $s$. The main runs use the secant slope.

\subsection{Offsets from $J$ Sub-intervals}
\label{app:s:subintervals}

\begin{remark}[Offsets from $J$ sub-intervals]
\label{rem:subintervals}
Cut $[\lo{z}, \up{z}]$ into $J$ equal sub-intervals $[a_k, b_k]$, $k = 1, \dots, J$. Let $e^{(k)}_i$ be the Bernstein coefficients of $\sigma(z) - s z$ on $[a_k, b_k]$ with the same slope $s$, that is, $e^{(k)}_i = c^{(k)}_i - s\,(a_k + (b_k - a_k) i/n)$, where $c^{(k)}$ are the Bernstein coefficients of $\sigma^{(r)}_j$ on $[a_k, b_k]$. Let $\up\beta_J$ and $\lo\beta_J$ be the largest and the smallest of all $e^{(k)}_i$. Then (i)~inequality~\eqref{eq:relaxation} holds with $\lo\beta_J, \up\beta_J$ in place of $\lo\beta, \up\beta$, and (ii)~$\lo\beta \le \lo\beta_J \le \up\beta_J \le \up\beta$. The case $J = 1$ is Lemma~\ref{lem:relaxation}. The implementation keeps the tighter of the two sets of offsets, which are both sound.
\end{remark}

\begin{proof}[Proof of Remark~\ref{rem:subintervals}]
(i)~Let $z \in [\lo{z}, \up{z}]$. Then $z$ lies in some sub-interval $[a_k, b_k]$. By (P4) and linearity, $e^{(k)}$ are the Bernstein coefficients of $p(z) = \sigma(z) - s z$ on $[a_k, b_k]$, and (P1) gives $p(z) \in [\min_i e^{(k)}_i, \max_i e^{(k)}_i]$. (ii)~The polynomial $p$ has Bernstein coefficients $e$ on $[\lo{z}, \up{z}]$. By (P2), its Bernstein coefficients $e^{(k)}$ on the sub-interval $[a_k, b_k]$ are convex combinations of $e_0, \dots, e_n$. Hence $\min_i e_i \le e^{(k)}_i \le \max_i e_i$ for all $k$ and $i$.
\end{proof}

\subsection{Quadratic Relaxation Width}
\label{app:s:band}

\begin{proposition}[Quadratic relaxation width]
\label{prop:band}
Let $n \ge 2$, let $[\lo{z}, \up{z}]$ have width $w = \up{z} - \lo{z}$, and let $\Gamma_2 \ge 0$ bound the absolute values of the Bernstein coefficients of $(\sigma^{(r)}_j)''$ on it. With the secant slope,
\begin{equation}
    \up\beta - \lo\beta \;\le\; \frac{n}{4(n-1)} \, \Gamma_2 \, w^2 .
\end{equation}
By item~(ii) of Remark~\ref{rem:subintervals}, the bound also holds for the sub-interval offsets.
\end{proposition}

\begin{proof}[Proof of Proposition~\ref{prop:band}]
The offsets are $\min_i e_i$ and $\max_i e_i$ by Lemma~\ref{lem:relaxation}. Let $p(s) = \sigma^{(r)}_j(\lo{z} + w s)$. Applying (P3) twice and the chain rule gives
\begin{equation*}
    p''(s) = n(n-1) \sum_i \Delta^2 c'_i\, B^{[0,1]}_{n-2,i}(s) = w^2 (\sigma^{(r)}_j)''(\lo{z} + w s),
\end{equation*}
where $\Delta^2 c'_i = c'_{i+2} - 2 c'_{i+1} + c'_i$. By uniqueness of the Bernstein form, $\Delta^2 c'_i = w^2 \zeta_i / (n(n-1))$, where $\zeta$ are the Bernstein coefficients of $(\sigma^{(r)}_j)''$ on the interval. Hence $|\Delta^2 c'_i| \le D := w^2 \Gamma_2 / (n(n-1))$. The linear term $s\,\xi_i$ does not change second differences, so $|\Delta^2 e_i| \le D$, and the secant slope gives $e_0 = e_n$. Set $u_i = e_i - e_0$ and $v_i = D\, i(n-i)/2$, so that $u_0 = u_n = v_0 = v_n = 0$ and $\Delta^2 v_i = -D$. The sequences $u + v$ and $v - u$ have second differences $\Delta^2 u_i - D \le 0$ and $-D - \Delta^2 u_i \le 0$. They are concave and vanish at both ends, so they are non-negative. Hence $|u_i| \le v_i \le n^2 D / 8$. Both the smallest and the largest $e_i$ lie within $n^2 D/8$ of $e_0$, so $\up\beta - \lo\beta \le n^2 D / 4 = n \Gamma_2 w^2 / (4(n-1))$.
\end{proof}

\subsection{Linear Bounds and Gap Identity}
\label{app:s:linear}

\begin{proof}[Proof of Theorem~\ref{thm:linear}]
For $\x \in \Rcal(P)$ we have $H f(\x) = \Lambda^{(R)} \sigma^{(R)}(\z^{(R)}(\x)) + o_{\mathrm{out}}$. Suppose that $H f(\x) \in [\Lambda^{(r)} \sigma^{(r)}(\z^{(r)}(\x)) + o_{\mathrm{lo}}^{(r)},\allowbreak\, \Lambda^{(r)} \sigma^{(r)}(\z^{(r)}(\x)) + o_{\mathrm{up}}^{(r)}]$ for constants $o^{(r)}_{\mathrm{lo}}, o^{(r)}_{\mathrm{up}}$. By assumption, $z^{(r)}_j(\x)$ lies in the interval on which neuron $(r,j)$ was relaxed, so~\eqref{eq:relaxation} gives $\sigma^{(r)}_j(z^{(r)}_j(\x)) \in [s^{(r)}_j z^{(r)}_j(\x) + \lo\beta^{(r)}_j, s^{(r)}_j z^{(r)}_j(\x) + \up\beta{}^{(r)}_j]$. Multiplying by the entries of $\Lambda^{(r)}$, the lower bound uses $\lo\beta$ for positive entries and $\up\beta$ for negative ones, and the upper bound the opposite, while both share the linear part $\Gamma^{(r)} \z^{(r)}(\x)$. Substituting $\z^{(r)}(\x) = W_r \sigma^{(r-1)}(\z^{(r-1)}(\x)) + \bbf_r$ yields the same statement at layer $r-1$ with coefficient $\Lambda^{(r-1)}$. At $r = 0$, $\sigma^{(0)}(\z^{(0)}(\x)) = \x$. Since both bounds share every $\Lambda^{(r)}$, they differ only in their constants, and $\cbf_{\mathrm{up}} - \cbf_{\mathrm{lo}} = \sum_r (\Lambda^{(r)+} - \Lambda^{(r)-}) \beta^{(r)} = \sum_r |\Lambda^{(r)}| \beta^{(r)}$.
\end{proof}

\section{Backward Contraction}
\label{app:proofs:phase2}

This appendix proves the statements of Section~\ref{sec:tightening}. It also states Lemma~\ref{lem:activation}, which that section cites.

\subsection{Backward Interval Propagation over One Layer}
\label{app:s:rows}

\begin{proof}[Proof of Lemma~\ref{lem:rows}]
Fix a row $j$ and $k$ with $W_{jk} \neq 0$. For $\sigma_{k'} \in [\lo{\sigma}_{k'}, \up{\sigma}_{k'}]$, the term $W_{jk'} \sigma_{k'}$ is at most $W^+_{jk'} \up{\sigma}_{k'} + W^-_{jk'} \lo{\sigma}_{k'}$ and at least $W^+_{jk'} \lo{\sigma}_{k'} + W^-_{jk'} \up{\sigma}_{k'}$, as in Lemma~\ref{lem:ibp}. Summing over $k' \neq k$ bounds the other terms of row $j$ by the two sums of the statement. Subtracting these sums and $b_j$ from $\lo{z}'_j \le \sum_{k'} W_{jk'} \sigma_{k'} + b_j \le \up{z}'_j$ gives the interval for $W_{jk} \sigma_k$. Dividing by $W_{jk}$ keeps the order of the ends when $W_{jk} > 0$ and reverses it when $W_{jk} < 0$. Every row holds for $\sigma$, so $\sigma_k$ lies in all these intervals and in $[\lo{\sigma}_k, \up{\sigma}_k]$. If their intersection is empty, no such $\sigma$ exists.
\end{proof}

\subsection{Monotonicity Certificate}
\label{app:s:clamped}

\begin{lemma}[Monotonicity certificate]
\label{lem:activation}\label{lem:clamped}
Let $[\lo{z}, \up{z}]$ be a bounded interval, and let $c'$ be the Bernstein coefficients of $\sigma^{(r)}_j$ on it. Then:
\begin{enumerate}
    \item $\sigma^{(r)}_j([\lo{z}, \up{z}]) \subseteq [\min_i c'_i, \max_i c'_i]$, and $c'_0 = \sigma^{(r)}_j(\lo{z})$, $c'_n = \sigma^{(r)}_j(\up{z})$.
    \item If $c'_{i+1} \ge c'_i$ for all $i$, then $\sigma^{(r)}_j$ is non-decreasing on $[\lo{z}, \up{z}]$ and its image is $[\sigma^{(r)}_j(\lo{z}), \sigma^{(r)}_j(\up{z})]$. Symmetrically, if $c'_{i+1} \le c'_i$ for all $i$, it is non-increasing and the image is $[\sigma^{(r)}_j(\up{z}), \sigma^{(r)}_j(\lo{z})]$.
\end{enumerate}
\end{lemma}

If item~2 applies, the neuron is \emph{certified monotone} on $[\lo{z}, \up{z}]$, more precisely certified non-decreasing or certified non-increasing. The test is sufficient only.

\begin{proof}[Proof of Lemma~\ref{lem:activation}]
Let $p(s) = \sigma^{(r)}_j(\lo{z} + (\up{z} - \lo{z}) s)$, whose Bernstein coefficients on $[0,1]$ are $c'$ by (P2). The enclosure and the end values $c'_0 = \sigma^{(r)}_j(\lo{z})$, $c'_n = \sigma^{(r)}_j(\up{z})$ are (P1), which proves item~1. 
By (P3),
\[
    p'(s) = n \sum_i (c'_{i+1} - c'_i)\, B^{[0,1]}_{n-1,i}(s),
\]
whose Bernstein coefficients $n(c'_{i+1} - c'_i)$ are non-negative if $c'_{i+1} \ge c'_i$ for all $i$. By (P1) applied to $p'$, $p' \ge 0$ on $[0,1]$, so $\sigma^{(r)}_j$ is non-decreasing on $[\lo{z}, \up{z}]$. A continuous non-decreasing function maps an interval onto the interval between its endpoint values. The non-increasing case follows by applying this to $-p$. This proves item~2.
\end{proof}

\subsection{Inverse of a Certified Monotone Activation}
\label{app:s:inverse}

\begin{proof}[Proof of Lemma~\ref{lem:inverse}]
The function $\sigma$ is continuous and non-decreasing on $[\lo{z}, \up{z}]$ by Lemma~\ref{lem:clamped}. It need not be strictly increasing. Let $S = \{ z \in [\lo{z}, \up{z}] : \sigma(z) \in [\lo{\sigma}', \up{\sigma}'] \}$.

\emph{Item~1.} If $\sigma(\up{z}) < \lo{\sigma}'$, then $\sigma(z) \le \sigma(\up{z}) < \lo{\sigma}'$ for every $z$, and if $\sigma(\lo{z}) > \up{\sigma}'$, then $\sigma(z) \ge \sigma(\lo{z}) > \up{\sigma}'$ for every $z$; in both cases $S = \emptyset$. Otherwise $\sigma(\lo{z}) \le \up{\sigma}'$ and $\sigma(\up{z}) \ge \lo{\sigma}'$. The set $\{z : \sigma(z) \ge \lo{\sigma}'\}$ is a closed upper set of the interval and $\{z : \sigma(z) \le \up{\sigma}'\}$ a closed lower set, so $S$ is their intersection, a closed interval, and it is non-empty because by continuity some $z$ has $\sigma(z) \in [\lo{\sigma}', \up{\sigma}']$.

\emph{Item~2.} If $\sigma(\lo{z}) \ge \lo{\sigma}'$, then $\lo{z}' = \lo{z}$ is below every element of $S$. Otherwise $\sigma(\lo{z}) < \lo{\sigma}' \le \sigma(\up{z})$, so the starting pair satisfies the property. A pair with $\sigma(a) < \lo{\sigma}' \le \sigma(b)$ has one half with the same property: $[a, m]$ if $\sigma(m) \ge \lo{\sigma}'$ and $[m, b]$ otherwise. Every $z \in S$ satisfies $\sigma(z) \ge \lo{\sigma}' > \sigma(a)$, hence $z > a$ because $\sigma$ is non-decreasing. So $\lo{z}' = a$ is below every element of $S$ after any number of halvings, and $a \ge \lo{z}$. The smallest element of $S$ lies in $(a, b]$, so its distance to $a$ is at most $b - a$, which halves at every step.

\emph{Item~3} is symmetric with the property $\sigma(a) \le \up{\sigma}' < \sigma(b)$: every $z \in S$ satisfies $\sigma(z) \le \up{\sigma}' < \sigma(b)$, hence $z < b$, and $b \le \up{z}$.

Both ends therefore satisfy $\lo{z} \le \lo{z}' \le z \le \up{z}' \le \up{z}$ for every $z \in S$, which is the claim. The non-increasing case follows by applying the lemma to $-\sigma$ and $[-\up{\sigma}', -\lo{\sigma}']$, since $\sigma(z) \in [\lo{\sigma}', \up{\sigma}']$ if and only if $-\sigma(z) \in [-\up{\sigma}', -\lo{\sigma}']$.
\end{proof}

\subsection{Soundness of Tightening}
\label{app:s:tighten}

\begin{proof}[Proof of Theorem~\ref{thm:tighten}]
Let $\x \in \Rcal(P) \cap \BRS_f(\Y)$. By Lemma~\ref{lem:ibp}, all its pre-activations and post-activations lie in the bounds of $\Bcal_P$, and $H f(\x) = H W_{R+1} \sigma^{(R)}(\z^{(R)}(\x)) + H \bbf_{R+1} \le h$. By Lemma~\ref{lem:rows}, $\sigma^{(R)}(\z^{(R)}(\x))$ lies in the contracted interval vector of line~1. Inductively, if $\sigma^{(r)}(\z^{(r)}(\x))$ lies in its contracted interval vector, then $z^{(r)}_j(\x)$ lies in the interval of Lemma~\ref{lem:inverse} for every certified neuron, and in its unchanged interval otherwise. Lemma~\ref{lem:rows} applied to $\z^{(r)}(\x) = W_r \sigma^{(r-1)}(\z^{(r-1)}(\x)) + \bbf_r$ then places $\sigma^{(r-1)}(\z^{(r-1)}(\x))$ in its contracted interval vector. Finally, $A \x + \cbf_{\mathrm{lo}} = L(\x) \le H f(\x) \le h$ by Theorem~\ref{thm:linear}, so the contraction of line~7 keeps $\x$. Hence $\x \in \Rcal(P')$, and no interval can become empty while such an $\x$ exists. The inclusion $\Rcal(P') \subseteq \Rcal(P)$ holds because every bound only shrinks.
\end{proof}

\subsection{Classification and Parts}
\label{app:s:polytopes}

\begin{proof}[Proof of Theorem~\ref{thm:polytopes}]
The row bound $\lo\rho_i$ ($\up\rho_i$) is the largest lower (smallest upper) bound on $(H f(\x))_i$ among the partial bounds of the back-substitution of Theorem~\ref{thm:linear}, each evaluated over the interval of its variable. Each partial bound is linear in a quantity that stays in its interval on $\Rcal(P)$ by Lemma~\ref{lem:ibp}, so the row bounds are valid on $\Rcal(P)$. Items~1 and~2 follow from Lemma~\ref{lem:ibp} and from this validity. For item~3, if $\x \in \Rcal(P)$ and $U(\x) \le h$, then $H f(\x) \le U(\x) \le h$ by Theorem~\ref{thm:linear}. For item~4, if $\x \in \Rcal(P) \cap \BRS_f(\Y)$, then $L(\x) \le H f(\x) \le h$. The polytope $\hat O(P)$ drops the constraints of layers $r \ge 2$ from $\Rcal(P)$, which can only enlarge the set. Items~3 and~4 also hold with $\Rcal(P)$ replaced by $\Rcal(P')$ for any pattern $P'$ with $\Rcal(P') \subseteq \Rcal(P)$, because the bounds of Theorem~\ref{thm:linear} hold on $\Rcal(P)$.
\end{proof}

\section{Branching}
\label{app:proofs:phase3}

This appendix gives the shares of the input score, the volume estimates, the initial partition and the parallel rounds of Section~\ref{sec:refinement}. It states and proves Propositions~\ref{prop:monotone} and~\ref{prop:concentration}, Lemma~\ref{lem:setdecomp} and Theorem~\ref{thm:soundness}, which that section cites.

\subsection{Shares of the Input Score}
\label{app:s:shares}

The input score of Section~\ref{sec:split} uses the shares
\begin{equation}
\begin{gathered}
    S^{(1)}_{ji} = \frac{|W_{1,ji}|\, h_i(\hat X')}{\sum_{i'} |W_{1,ji'}|\, h_{i'}(\hat X')}, \qquad S^{(r)} = N^{(r)} S^{(r-1)}, \\
    N^{(r)}_{jj'} = \frac{|W_{r,jj'}| \,(\up{\sigma}^{(r-1)}_{j'} - \lo{\sigma}^{(r-1)}_{j'})}{\sum_{j''} |W_{r,jj''}| \,(\up{\sigma}^{(r-1)}_{j''} - \lo{\sigma}^{(r-1)}_{j''})},
\end{gathered}
\end{equation}
where $S^{(r)}_{ji}$ is the share of the width of $z^{(r)}_j$ that is due to $x_i$, and $N^{(r)}$ is the matrix of the shares of the post-activation widths of layer $r-1$ in the width of each pre-activation of layer $r$. Every row of $S^{(1)}$ and of $N^{(r)}$ sums to one, so for each neuron the shares $S^{(r)}_{ji}$ of all input dimensions sum to one. With $S^{(r)}_{ji} = 1$ the input score of a neuron equals its neuron score $\tfrac34\theta_{rj}$.

\subsection{Monotone Splits}
\label{app:s:monotone}

\begin{proposition}[Monotone splits]
\label{prop:monotone}
Unless it is constant, the activation $\sigma^{(r)}_j$ has at most $n - 1$ critical points. Suppose that every critical point of $\sigma^{(r)}_j$ in $[\lo{z}, \up{z}]$ is simple. Then for every sufficiently fine subdivision of $[\lo{z}, \up{z}]$ whose split points include these critical points, the monotonicity certificate of Lemma~\ref{lem:activation} holds on every piece.
\end{proposition}

The proposition is a statement about the split points. It does not claim that the selection rule produces a fine subdivision.

\begin{proof}[Proof of Proposition~\ref{prop:monotone}]
The polynomial $(\sigma^{(r)}_j)'$ has degree $n - 1$, so it has at most $n-1$ roots unless it is identically zero. If it is identically zero, $\sigma^{(r)}_j$ is constant and all Bernstein coefficients are equal. Consider a piece $[c, c + w]$ of the subdivision.

Suppose that the piece starts at a simple root $c$ of $(\sigma^{(r)}_j)'$. The Bernstein coefficients of $(\sigma^{(r)}_j)'$ on the piece equal its values at the grid points $c + w\, i/(n-1)$ up to $\mathcal{O}(w^2)$, and these values are $(\sigma^{(r)}_j)''(c)\, w\, i/(n-1) + \mathcal{O}(w^2)$. The first Bernstein coefficient equals $(\sigma^{(r)}_j)'(c) = 0$. Since $(\sigma^{(r)}_j)''(c) \neq 0$, all other Bernstein coefficients have the sign of $(\sigma^{(r)}_j)''(c)$ for small $w$. By (P3), the differences of the Bernstein coefficients of $\sigma^{(r)}_j$ on the piece are proportional to the Bernstein coefficients of $(\sigma^{(r)}_j)'$, so they do not change sign.

A piece that ends at a root is symmetric. A piece that contains no root has $|(\sigma^{(r)}_j)'|$ bounded away from zero, and the Bernstein coefficients of $(\sigma^{(r)}_j)'$ on a fine piece converge to its values uniformly, so they share one sign. For a sufficiently fine subdivision, each piece falls into one of these cases.
\end{proof}

\subsection{Volume Estimates and Queue Order}
\label{app:s:mass}

The estimated volume of a pattern is the estimated share of the input box $\X$ that its region covers. Let $n_{\mathrm{in}}$ of the $M$ samples of $\hat X'$ lie in $\Rcal(P')$ and $n_C$ in the region of child $C$. The estimated volumes relative to $\X$ are
\begin{equation}
    \hat m(P') = \frac{\vol(\hat X')}{\vol(\X)} \cdot \frac{\max(n_{\mathrm{in}}, 1)}{M + 1}, \qquad \hat m(C) = \hat m(P') \cdot \frac{n_C + \tfrac12}{n_{\mathrm{in}} + 1}.
    \label{eq:mass}
\end{equation}
A tail of a joint split receives the Gaussian tail probability $\eta = \Phi(-\kappa) \approx 3.2\cdot 10^{-5}$ of its parent, with $\Phi$ the standard normal distribution function, and the central part receives $1 - 2\eta$. These values replace the counts of~\eqref{eq:mass}, so the children~\eqref{eq:setdec} have the volumes
\begin{equation}
    \hat m(T_t^{\pm}) = \hat m(P')\, \eta\, (1 - 2\eta)^{t-1}, \qquad \hat m(K) = \hat m(P')\, (1 - 2\eta)^{|S|}.
    \label{eq:massset}
\end{equation}
The patterns of the initial partition have these volumes with $\hat m(P') = 1$. The estimates are heuristic and only order the priority queue. A round takes at most $B$ of the pending patterns of largest estimated volume, and only those with estimated volume at least $\gamma = 10^{-3}$ times the largest. The pattern of largest estimated volume is always taken.

\subsection{Concentration}
\label{app:s:concentration}

\begin{proposition}[Concentration]
\label{prop:concentration}
Let $\x$ be uniform on a box $\hat X$, let $z = \wbf^\top \x + b$, and let $v_i = |w_i|\, h_i(\hat X)$. The interval of $z$ over $\hat X$ has width $\|\vbf\|_1$, and $z$ has standard deviation $\|\vbf\|_2 / \sqrt{12}$. Their ratio lies in $[\sqrt{12}, \sqrt{12 d}]$.
\end{proposition}

\begin{proof}[Proof of Proposition~\ref{prop:concentration}]
The coordinates are independent and uniform, so the variance of $w_i x_i$ is $v_i^2/12$, and the variance of $z$ is $\|\vbf\|_2^2/12$. The interval width is $\sum_i v_i = \|\vbf\|_1$. The bounds follow from $\|\vbf\|_2 \le \|\vbf\|_1 \le \sqrt{d}\, \|\vbf\|_2$.
\end{proof}

Eligibility is typical in high dimension. For weights with i.i.d.\ Gaussian entries and equal edge lengths, $\|\vbf\|_1 / \|\vbf\|_2 \approx \sqrt{2d/\pi}$. The ratio of Proposition~\ref{prop:concentration} is then about $2.76\sqrt{d}$. By the central limit theorem, $z$ is approximately Gaussian. The interval $[\mu - 4\varsigma, \mu + 4\varsigma]$, with mean $\mu$ and standard deviation $\varsigma$, holds all but about $6 \cdot 10^{-5}$ of the volume. It covers a fraction of about $2.9/\sqrt{d}$ of the interval of $z$, which is about $0.09$ for $d = 1024$. A bisection only halves such an interval, so its children stay much wider than the part that holds the volume. A restriction of $z$ to $[\mu - 4\varsigma, \mu + 4\varsigma]$ shrinks the interval by a factor of about $2.9/\sqrt{d}$ and moves almost no volume into the tails. For a first-layer neuron and up to sampling error, the proposition shows that eligibility is possible only if $\sqrt{12 d} > 2\kappa/\omega$, that is $d > 21$ for $\kappa = 4$ and $\omega = 1/2$. Deeper neurons are not covered by the proposition, and the test is applied to them as well.

\subsection{Joint Split}
\label{app:s:setdecomp}

\begin{lemma}[Joint split]
\label{lem:setdecomp}
The at most $2|S| + 1$ children~\eqref{eq:setdec} cover $\Rcal(P')$. Two different children intersect only in a set on which some pre-activation $z_t$ equals an endpoint of $C_t$.
\end{lemma}

\begin{proof}[Proof of Lemma~\ref{lem:setdecomp}]
Let $\x \in \Rcal(P')$. If $z_t(\x) \in C_t$ for all $t$, then $\x \in K$. Otherwise let $t$ be the smallest index with $z_t(\x) \notin C_t$. Then $z_s(\x) \in C_s$ for $s < t$, and $z_t(\x) < \underline c_t$ or $z_t(\x) > \overline c_t$, so $\x \in T_t^-$ or $\x \in T_t^+$. An omitted tail adds nothing: if $\underline c_t$ equals the lower end of the interval of neuron $t$, every input of $\Rcal(P')$ with $z_t \le \underline c_t$ has $z_t = \underline c_t \in C_t$ and lies in a later child, and the same holds at the upper end. For the second claim, let two different children carry the indices $t < t'$, where $K$ has the index $|S|+1$. The child of index $t'$ requires $z_t \in C_t$. The child of index $t$ requires $z_t \le \underline c_t$ or $z_t \ge \overline c_t$. Both hold only if $z_t$ equals an endpoint of $C_t$. The two tails of the same index meet only where $\underline c_t = \overline c_t$, which is also an endpoint.
\end{proof}

\subsection{Initial Partition}
\label{app:s:rootcover}

Before the first round, \tool replaces the root by a joint split of the first layer, with closed-form statistics instead of samples. For uniform inputs on $\X$, the mean $\mu_j$ of $z^{(1)}_j$ is its value at the centre of $\X$, and its variance is $\varsigma_j^2 = \sum_i W_{1,ji}^2 h_i(\X)^2/12$ (Proposition~\ref{prop:concentration}). Let $j_1, \dots, j_p$ be the neurons whose central interval $[\mu_j - \kappa\varsigma_j, \mu_j + \kappa\varsigma_j]$ is narrower than $\omega$ times the range $\sum_i |W_{1,ji}| h_i(\X)$ of $z^{(1)}_j$ over $\X$. The joint split~\eqref{eq:setdec} with this set gives $2p + 1$ patterns that cover the root, the \emph{initial partition}. If $p = 0$, the root is kept.

\subsection{Parallel Rounds, Backends and Budget}
\label{app:s:parallel}

The patterns of a round are processed independently of each other. The batched PyTorch backend processes a round as one pass on the device, and the NumPy backend processes its patterns one by one. With the same samples, the two backends make the same decisions up to floating-point summation. In the reported PyTorch runs the samples come from the PyTorch generator, so the search trees differ from those of the NumPy backend. Algorithm~\ref{alg:overview} processes at most $N$ patterns with splitting and then classifies the pending ones without splitting, in the order of largest estimated volume first.

\subsection{Soundness of \tool}
\label{app:s:soundness}

The root pattern has $\hat X = \X$ and intervals $\hat Z^{(r)}$ that enclose $\z^{(r)}(\X)$, so its region is $\X$. \textsc{Collect} adds an accepted region to both $\Ical$ and $\Ocal$, discards a pruned pattern, inserts the children of a split into $\Qcal$, and adds the parts $I$ and $O$ of an undecided pattern to $\Ical$ and $\Ocal$.

\begin{theorem}[Soundness of \tool]
\label{thm:soundness}
\tool terminates. It processes at most $N + (2p+1) + 2 W_\Sigma N$ patterns, and at most $2N$ if the final classification is limited to $N$ patterns, where $2p+1$ is the size of the initial partition (Appendix~\ref{app:s:rootcover}) and $W_\Sigma$ the number of hidden neurons. At the start of every round,
\begin{equation}
    \bigcup \Ical \;\subseteq\; \BRS_f(\Y) \;\subseteq\; \bigcup \Ocal \,\cup\, \bigcup_{P \in \Qcal} \Rcal(P),
    \label{eq:invariant}
\end{equation}
where $\Qcal$ is the set of pending patterns, and the returned families satisfy~\eqref{eq:problem}.
\end{theorem}

Stopping after any round and adding the pending regions to $\Ocal$ therefore yields a valid answer. Every step is polynomial in the input dimension, the widths, the depth and the degree. The cost that is exponential in the worst case is confined to the number of patterns.

\begin{proof}[Proof of Theorem~\ref{thm:soundness}]
\emph{Inner inclusion.} Sets are added to $\Ical$ only as accepted regions or as inner parts. Both are contained in $\BRS_f(\Y)$ by Theorem~\ref{thm:polytopes}, applied to the pattern for which the set was produced. Explicit accepted regions are accepted on a superset of the pattern region, so the same theorem applies to them.

\emph{Outer inclusion.} Initially $\Qcal$ holds the root, whose region is $\X \supseteq \BRS_f(\Y)$. With the initial partition, $\Qcal$ holds $2p + 1$ patterns that cover $\X$ (Appendix~\ref{app:s:rootcover} and Lemma~\ref{lem:setdecomp}). When a pattern $P$ is processed, the points of $\Rcal(P) \cap \BRS_f(\Y)$ remain covered. If $P$ is pruned, there are none (Theorems~\ref{thm:tighten} and~\ref{thm:polytopes}). If it is accepted or undecided, they lie in the set added to $\Ocal$, by Theorem~\ref{thm:polytopes} applied to the contracted pattern $P'$ and by $\Rcal(P) \cap \BRS_f(\Y) \subseteq \Rcal(P')$ from Theorem~\ref{thm:tighten}. A pattern whose bounds are not finite is returned as its own region, which contains these points. If $P$ is split, its children cover $\Rcal(P')$. A neuron split covers the split interval with two intervals. A joint split covers $\Rcal(P')$ with at most $2|S| + 1$ children (Lemma~\ref{lem:setdecomp}). An input split covers the box with two boxes. All other constraints are those of $P'$.

\emph{Termination.} Every round of the refinement stage processes at least one pattern, because the pending pattern of largest estimated volume always passes the volume gate (Appendix~\ref{app:s:mass}), and the counter of processed patterns never exceeds $N$ in this stage. The stage therefore ends after at most $N$ processed patterns. Each processed pattern is replaced by at most $2W_\Sigma + 1$ children, since $|S| \le W_\Sigma$. At most $2p + 1 + 2 W_\Sigma N$ patterns are pending at the end of the stage. The final classification processes pending patterns without splitting and creates none, so it ends after a finite number of rounds. If it is limited to $N$ patterns, the remaining ones are added to $\Ocal$ as their regions. Then $\Qcal = \emptyset$, and~\eqref{eq:invariant} is~\eqref{eq:problem}.
\end{proof}

\section{Settings of the Runs}
\label{app:exp:setup}

Table~\ref{tab:experiments} lists the experiments and the case studies. The ratios of Section~\ref{sec:experiments} are defined for $\vol(\BRS_f(\Y)) > 0$. The volumes are estimated on $20{,}000$ uniform samples of the input box. Every run is checked on these samples. No sampled input of the inner set lies outside the BRS, and no sampled input of the BRS lies outside the outer set. All runs use the defaults of Sections~\ref{sec:bounds} to~\ref{sec:refinement} with $M = 4{,}096$ volume samples, $N = 400$ unless stated otherwise, and every pending pattern classified at the end. The GPU is one NVIDIA A30, and TinyBERT and the two-input networks run on CPU cores.

\begin{table}[h]
\caption{The experiments and the case studies. Trained networks have degree $9$. A run is one execution of \tool at one budget.}
\label{tab:experiments}
\footnotesize\setlength{\tabcolsep}{2.5pt}\renewcommand{\arraystretch}{1.05}
\noindent\begin{tabular}{@{}>{\raggedright\arraybackslash}p{0.21\columnwidth}>{\raggedright\arraybackslash}p{0.31\columnwidth}>{\raggedleft\arraybackslash}p{0.09\columnwidth}>{\raggedright\arraybackslash}p{0.225\columnwidth}>{\raggedleft\arraybackslash}p{0.08\columnwidth}@{}}
\toprule
Experiment & Network & Inputs & Training & Runs \\
\midrule
2-D, random & MLP, up to width 500, depth 6, degree 15 & 2 & none & 859 \\
Images & MLP, width 100--500, depth 2--4 & 784; 3,072 & Appendix~\ref{app:exp:images} & 108 \\
Case study 1, closed loop & MLP, width 128, depth 1, merged up to 40 layers & 17 & Appendix~\ref{app:exp:halfcheetah} & 200 \\
Case study 2, transformer & Last block of TinyBERT, 3 heads & 312 & Appendix~\ref{app:exp:tinybert} & 43 \\
\bottomrule
\end{tabular}
\end{table}

\section{Image Classifiers}
\label{app:exp:images}
\label{app:exp:images:training}

This appendix gives the training of Bernstein networks~1 and~2 of Experiment 2. On MNIST~\cite{lecun1998mnist}, GTSRB~\cite{stallkamp2012gtsrb} and CIFAR-10~\cite{krizhevsky2009}, inputs are flattened and scaled to $[0, 1]$, without data augmentation. GTSRB images are cropped to the sign and resized to $32 \times 32$. The networks have no batch normalization. Every hidden neuron has its own learned Bernstein activation of degree $n = 9$ on the bounds $[-5, 5]$.

Both networks use AdamW with weight decay $10^{-4}$, a one-cycle learning rate with peak $10^{-3}$ and 10\,\% warm-up, 40 epochs, batch size 128 and seed 0. Before the first step, each linear layer is rescaled so that its pre-activations have standard deviation 1.5 on 2,000 training images. The penalty $5 \sum_{r} \operatorname{mean}_{j} \max(|z_{rj}| - 5, 0)^2$ on the pre-activations $z_{rj}$ of hidden layer $r$ keeps the training data inside the bounds.

Bernstein network~1 uses bi-Lipschitz training with the cross-entropy loss. Each square weight matrix, between hidden layers, is parametrized as $W = \operatorname{diag}(s)\, Q$, with $Q$ orthogonal and $s_j = g \exp(\tfrac12 \ln 4 \, \tanh t_j)$, so its condition number is at most 4. Neighbouring Bernstein coefficients satisfy $c_{k+1} - c_k \in [0.25/n, 4/n]$, so every activation is increasing, with slope between 0.025 and 0.4. The weight matrices from the inputs and to the outputs stay free.

Bernstein network~2 uses certified training. It minimizes $\alpha L_{\mathrm{clean}} + (1 - \alpha) L_{\mathrm{robust}}$ as in DeepBern-Nets~\cite{khedr2024deepbern}, with the penalty. The term $L_{\mathrm{robust}}$ is the cross-entropy of the worst-case logits, from interval bound propagation~\cite{gowal2018ibp} over the box of radius $\varepsilon$ around each training input. Once the robust term starts, $\alpha$ falls linearly from 0.9 towards 0 over 14 epochs, down to its floor. Table~\ref{tab:images:recipes} gives the rule for $\varepsilon$. A ramp raises it linearly to its target over the same 14 epochs. The adaptive rule starts at one tenth of the target and, after each epoch, multiplies $\varepsilon$ by 1.25 if the certified accuracy on the first 2,000 test images is at least 0.8, and by 0.8 otherwise, between one hundredth of the target and the target.

\begin{table}[t]
\centering
\caption{Recipes of Bernstein network~2. Floor: the smallest $\alpha$. Robust from: the first epoch with a robust loss. Kept from: the first epoch that may be kept as the final network.}
\label{tab:images:recipes}
\footnotesize\setlength{\tabcolsep}{3pt}\resizebox{\columnwidth}{!}{\begin{tabular}{lrrlrrrr}
\toprule
Data set & Width & Depth & Rule & Target $\varepsilon$ & Floor & Robust from & Kept from \\
\midrule
MNIST & 100 & 2 to 4 & ramp & 0.01 & 0 & 2 & 16 \\
MNIST & 300 & 4 & ramp & 0.002 & 0.5 & 2 & 16 \\
MNIST & 500 & 4 & ramp & 0.001 & 0.5 & 2 & 16 \\
GTSRB, CIFAR-10 & 100 & 2 to 4 & adaptive & 0.004 & 0.5 & 7 & 21 \\
GTSRB, CIFAR-10 & 300 & 4 & adaptive & 0.002 & 0.5 & 7 & 21 \\
GTSRB, CIFAR-10 & 500 & 4 & adaptive & 0.001 & 0.5 & 7 & 21 \\
\bottomrule
\end{tabular}
}
\end{table}

Every network keeps the epoch with the best test accuracy among the eligible ones, given in Table~\ref{tab:images:recipes} for Bernstein network~2. For Bernstein network~1 these are all epochs on MNIST, except from epoch 16 at width 500, and the epochs from 16 on for GTSRB and CIFAR-10. The selection uses the test set, so the test accuracies are slightly optimistic.

\section{Case Study 1: A Hybrid Dynamical Control System}
\label{app:exp:halfcheetah}

This appendix gives the exact merge of the loop into the network form, the training of the controller, the plant, the runs and the settings of Case Study 1, and frames of one fall.

\subsection{Exact Merge of the Closed Loop}
\label{app:s:loop}\label{app:benchmarks:loop}

The closed loop is merged, without training, into the form of Definition~\ref{def:bernn}. The merged network equals the loop on the input box, so its certificates hold for the loop. Let $x_t \in \R^{d}$ be the state. A Bernstein controller $\pi$ with $R_\pi$ hidden layers acts on an affine, step-dependent plant (for example a linearization along a nominal trajectory),
\begin{equation}
    x_{t+1} = F_t(x_t) = A_t x_t + B_t \clip\big(\pi(x_t), -1, 1\big) + c_t ,
    \label{eq:loop}
\end{equation}
where $A_t$, $B_t$, $c_t$ are the matrices and offset of step $t$. For a horizon $K$ the $K$-step backward reachable set is $\BRS^K(\Y) = \{\x_0 \in \X : F_{K-1} \circ \dots \circ F_0 (\x_0) \in \Y\}$, with the target $\Y$ on the last state.

In the merge, each hidden layer holds the controller's neurons and $d$ copy neurons that carry the state. A clip layer of identity neurons (Lemma~\ref{lem:identity}) with bounds $[-1,1]$ follows $\pi$ and represents the action limits of the plant, and each plant step is one affine map composed with the next controller layer. One step uses $R_\pi + 1$ activation layers, and the merged network $f_K$ maps $\x_0$ to $x_K$. The copy neurons receive their bounds from interval bounds over $\X$, widened on each side by the width of the interval.

\begin{lemma}[Identity neuron]
\label{lem:identity}
Let neuron $(r,j)$ have bounds $[\ell, u]$, $\ell < u$, and Bernstein coefficients
\begin{equation}
    c_i = \ell + (u - \ell)\, i/n, \quad i = 0, \dots, n.
    \label{eq:identity_cp}
\end{equation}
Then $\sigma^{(r)}_j(z) = z$ on $[\ell, u]$. On a pattern with pre-activation interval $[\lo z, \up z] \subseteq [\ell, u]$, the neuron is certified non-decreasing, its secant slope is $1$, its relaxation width is $\beta^{(r)}_j = 0$, its contribution to every gap is $0$ so that branching never splits it, and its inverse on the interval is exact.
\end{lemma}

\begin{proof}[Proof of Lemma~\ref{lem:identity}]
By (P4) the polynomial with Bernstein coefficients \eqref{eq:identity_cp} on $[\ell,u]$ is $z \mapsto z$. On $[\lo z,\up z]$ the Bernstein coefficients are $c'_i = \lo z + (\up z - \lo z)\, i/n$ by (P2) and (P4). They are non-decreasing, so the neuron is certified non-decreasing by Lemma~\ref{lem:activation}. The secant slope is $(c'_n - c'_0)/(\up z - \lo z) = 1$, and the offsets of Lemma~\ref{lem:relaxation} are $e_i = c'_i - \lo z - (\up z - \lo z)\, i/n = 0$, so its relaxation width is $0$. By the gap identity~\eqref{eq:gap_identity}, the contribution of a neuron to the gap of a row is the magnitude of its back-substituted coefficient times its relaxation width, so it is $0$. A neuron split is scored by this contribution (Section~\ref{sec:split}), so branching never splits it. The inverse is exact because $\sigma^{(r)}_j(z) = z$ on the interval.
\end{proof}

\begin{proposition}[Exact merge of the closed loop]
\label{prop:loop}
Let every copy neuron have bounds containing the interval bounds over $\X$ of the coordinate it carries, and let every action stay within $[-1, 1]$ on $\X$. Then:
\begin{enumerate}
    \item $f_K(\x_0) = F_{K-1} \circ \dots \circ F_0(\x_0)$ for every $\x_0 \in \X$, hence $\{\x_0 \in \X : f_K(\x_0) \in \Y\} = \BRS^K(\Y)$;
    \item every copy neuron has relaxation width $0$ on each pattern inside its bounds, so it is never split.
\end{enumerate}
\end{proposition}

The guarantee covers the loop given to the solver.

\begin{proof}[Proof of Proposition~\ref{prop:loop}]
(1) Copy neurons output their input for every $\x_0 \in \X$ (Lemma~\ref{lem:identity}). The controller's neurons receive the same inputs as in $\pi$. The clip layer outputs the action, which equals $\clip(\pi(x_t),-1,1)$ because the action stays within $[-1, 1]$. The composed affine map equals the plant step followed by the first controller layer, since composition of affine maps is affine and associative. Induction over $t$ gives $f_K = F_{K-1} \circ \dots \circ F_0$. (2) This follows from Lemma~\ref{lem:identity}.
\end{proof}

\subsection{The Controller and Its Training}
\label{app:exp:halfcheetah:controller}

The verified network is the trained controller itself, and no teacher is distilled into it. It has 17 inputs, one hidden layer of 128 units, 6 outputs and Bernstein activations of degree 9. Proximal policy optimization~\cite{schulman2017ppo} in Stable-Baselines3~\cite{raffin2021sb3} trains it on the HalfCheetah-v4 environment of Gymnasium~\cite{towers2024gymnasium}, with Bernstein layers in place of hyperbolic tangent layers, for one million steps with one seed. The actor outputs the mean action without a squashing function, and the simulator clips the action to $[-1, 1]$, which is the clip layer of the merged network. The critic is not verified and keeps the hyperbolic tangent network of the library.

We adapt the certified training of DeepBern-Nets~\cite{khedr2024deepbern} to PPO. That training uses the cross-entropy of the worst-case logits, which needs class labels. PPO has no labels, so our certified training adds a term on the width of the action interval to the loss of the library,
\begin{equation*}
L = L_{\mathrm{PPO}} + \lambda_t \, \frac{1}{6 |\mathcal{M}|} \sum_{o \in \mathcal{M}} \sum_{j=1}^{6} \big( \up{u}_j(o) - \lo{u}_j(o) \big)^2 .
\end{equation*}
Here $\mathcal{M}$ is the mini-batch, and $[\lo{u}_j(o), \up{u}_j(o)]$ is the interval of the mean action $j$ over the box of radius $\eps_t$ around the normalized observation $o$, computed by interval bound propagation. The radius $\eps_t$ and the weight $\lambda_t$ ramp linearly from zero to their final values of 0.05 and 3 between 10\% and 60\% of the steps. The radius is in standardized observation units.

The running observation normalization is folded into the first affine layer at export, so that the exported network takes raw states. During training the bounds of each neuron follow the observed range of its pre-activation, which only grows, widened by 10\,\% of that range on each side. The export only widens them, to contain the pre-activations on the last 100,000 training observations and on 20 fresh episodes, plus 10\,\% of their range on each side. In ten further fresh deterministic episodes, every pre-activation lies inside the bounds of its neuron.

The training uses a learning rate of $10^{-4}$, rollouts of 512 steps, mini-batches of 64, 20 epochs per update, a discount of 0.98, a GAE parameter of 0.92, a clip range of 0.1, an entropy coefficient of $4.02 \times 10^{-4}$, a value coefficient of 0.581, a maximum gradient norm of 0.8 and an initial $\log \sigma$ of $-2$. These settings start from the published configuration for HalfCheetah of the RL Baselines3 Zoo~\cite{raffin2020zoo}, and a sweep of nineteen runs sets the learning rate and the final weight. The versions are Stable-Baselines3 2.3.2, Gymnasium 0.29.1 and MuJoCo 3.1.6. A controller has no test or certified accuracy, so we report its return, which is $4{,}339 \pm 386$ on 10 fresh episodes. Its action half-width is 0.0455 at the final radius.

\begin{figure*}[t]
\centering
\includegraphics{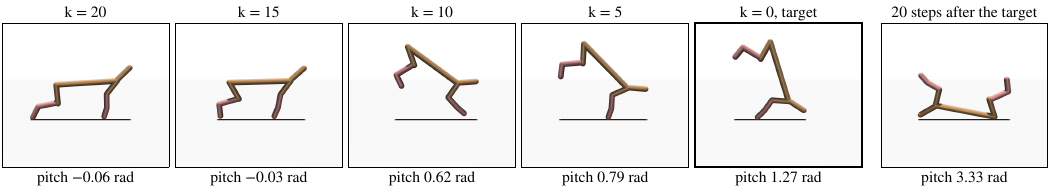}
\caption{Frames of the logged fall of episode~11, with the pitch below each frame. One step is 0.05\,s, and the camera follows the robot. The frames run from $k = 20$ to the target step $E$, and the last frame lies past the target.}
\label{fig:halfcheetah:frames}
\end{figure*}

\subsection{Plant, Runs and Settings}
\label{app:exp:halfcheetah:settings}

Central finite differences of the whole simulator step give $A_t$, $B_t$ and $c_t$ of~\eqref{eq:loop} along a closed-loop path. With $R_\pi = 1$, one step gives 2 activation layers, so the merged network has $2k$ activation layers, with 40 layers and 3,360 units on 17 inputs at $k = 20$. Each trajectory has one linearization from step $E - 20$, and the networks of one trajectory share their tails.

Table~\ref{tab:halfcheetah:trajectories} lists the five falls that we analyze. The linearization path of episode~49 departs from the logged states by up to 1.6 half-widths of the thin box, while the other four stay within 0.04.

\begin{table}[htbp]
\centering
\caption{The five falls: the episode, the target step $E$, the twenty start steps and the logged pitch at $E$.}
\label{tab:halfcheetah:trajectories}
\small
\begin{tabular}{rrrr}
\toprule
episode & target step $E$ & start steps & logged pitch at $E$ (rad) \\
\midrule
11 & 450 & 430 to 449 & 1.27 \\
24 & 504 & 484 to 503 & 1.43 \\
25 & 254 & 234 to 253 & 1.39 \\
23 & 491 & 471 to 490 & 1.10 \\
49 & 350 & 330 to 349 & 1.51 \\
\bottomrule
\end{tabular}

\end{table}

Each run takes a box around the logged state at its start step, with the half-widths of Table~\ref{tab:halfcheetah:boxes}, where $S_m$ is $m$ times the base half-width $S_1$. The base half-widths are representative sensor errors, chosen and not measured on this simulator. Table~\ref{tab:halfcheetah:scales} gives the scale of each coordinate in the wide box and in the thin box. No error enters the loop at later steps.

\begin{table}[htbp]
\centering
\caption{Half-widths per group of state coordinates, rounded within each group.}
\label{tab:halfcheetah:boxes}
\small
\begin{tabular}{llrrr}
\toprule
coordinates & unit & $S_1$ & $S_4$ & $S_{16}$ \\
\midrule
height & m & 0.005 & 0.02 & 0.08 \\
pitch & rad & 0.01 & 0.04 & 0.16 \\
six joint angles & rad & 0.002 & 0.008 & 0.032 \\
forward speed & m/s & 0.1 & 0.4 & 1.6 \\
vertical speed & m/s & 0.1 & 0.4 & 1.6 \\
pitch rate & rad/s & 0.05 & 0.2 & 0.8 \\
six joint speeds & rad/s & 0.3 & 1.2 & 4.8 \\
\bottomrule
\end{tabular}

\end{table}

\begin{table}[htbp]
\centering
\caption{The scale of each group of state coordinates in the two boxes, with the half-widths of Table~\ref{tab:halfcheetah:boxes}. Bold: the coordinates at $S_{16}$ in both boxes.}
\label{tab:halfcheetah:scales}
\small
\begin{tabular}{lcc}
\toprule
coordinates & wide box & thin box \\
\midrule
height & $S_{16}$ & $S_4$ \\
\textbf{pitch} & \textbf{\boldmath $S_{16}$} & \textbf{\boldmath $S_{16}$} \\
six joint angles & $S_{16}$ & $S_4$ \\
\textbf{forward speed} & \textbf{\boldmath $S_{16}$} & \textbf{\boldmath $S_{16}$} \\
vertical speed & $S_{16}$ & $S_4$ \\
\textbf{pitch rate} & \textbf{\boldmath $S_{16}$} & \textbf{\boldmath $S_{16}$} \\
six joint speeds & $S_{16}$ & $S_4$ \\
\bottomrule
\end{tabular}

\end{table}

Algorithm~\ref{alg:overview} processes at most $N = 2{,}000$ patterns with splitting and then classifies every pending pattern, so the number of processed patterns can exceed $2N$. Its median is 3,898 and its maximum is 20,211. The other settings are $B = 64$, $J = 2$, $M = 4{,}096$, $\varphi = 0.8$, $\kappa = 4$, $\omega = 0.5$ and the initial partition, in PyTorch with double precision on one A30 GPU per job. The shares use 20,000 samples per box. All runs took 12.4 GPU hours. Input splits make up 99.6\% of the splits, and joint splits occur in 19 of the 200 runs.

All 200 runs pass the soundness check, for \tool and for the baseline: no sampled state of an inner set lies outside the true set, and no sampled state of the true set lies outside an outer set. Copy neurons never leave their bounds in any run. On 500 box points per run, the merged network and the loop differ by at most 0.021, where states reach $10^{6}$, a relative difference of $2 \times 10^{-8}$. This is consistent with floating-point error amplified by the loop, and it was not tested further.

The baseline is CROWN with guided input partitioning, on the same merged networks, targets, boxes and samples. Its element budget doubles from one element up to two caps, 262,144 elements and 1,800 s per box. Each box uses the last baseline run within the wall-clock time of \tool on that box. On 6 boxes with $k \le 3$ even the run of one element takes longer than \tool, and that run is used. Doubling leaves the baseline a median of 0.57 of the time of \tool. The first run past that time adds 0.007 to the inner share on average and overtakes \tool on none of 185 boxes. The cap of elements applies on 8 boxes and the cap of time on 1.

\subsection{One Logged Fall}
\label{app:exp:halfcheetah:figures}

Figure~\ref{fig:halfcheetah:frames} shows the fall of episode~11 in MuJoCo. The robot tips forward over its head and lands on its back, and it stays there for the remaining 530 steps of the run.

\section{Case Study 2: Transformer}
\label{app:exp:tinybert}

This appendix gives the training of the student  network used in Case Study 2 and the setup of the runs.

\subsection{Training}
\label{app:exp:tinybert:networks}

The teacher is TinyBERT~\cite{jiao2020tinybert}, a distilled BERT~\cite{devlin2019bert} with 4 blocks of width 312 and 12 attention heads. We fine-tune it on SST-2~\cite{socher2013sst,wang2019glue} for 5 epochs with AdamW. The 872 development sentences serve as the test set, and the teacher reaches 88.65\,\% on them.

The student replaces the last block, the pooler and the classifier, while the first three blocks stay frozen. It has three heads of width 26 with a monotone Bernstein gate of degree 9 in place of the softmax, an MLP of 300 Bernstein curves of degree 9, and the pooler's dense map with 312 Bernstein curves of degree 9 in place of the tanh. The classifier reads only the [CLS] row, so the student computes only that row.

The student is initialized from the teacher. Its loss is the distillation loss of Hinton et al.~\cite{hinton2015} with the cross-entropy, a match of the [CLS] output, and a penalty that keeps each pre-activation inside the domain of its curve. Training runs for 30 epochs with AdamW, and we keep the epoch with the best agreement with the teacher on held-out sentences.

The student reaches 88.53\,\% on the development sentences.

\subsection{Setup}
\label{app:exp:tinybert:setup}

The free word of a sentence is the content word with the largest attention from the [CLS] token in the teacher's last block. With the other words fixed, the student is rewritten into a Bernstein network. One word is free, so each attention score is affine in the word's point and needs no squares. Each product of a gate output and a value becomes a difference of two squares, and identity neurons (Lemma~\ref{lem:identity}) carry the free word's point and the residual stream. The rewritten network has four layers on 312 inputs, with 1,395 neurons in total, and it agrees with the student to at most $5.3\cdot 10^{-15}$ on 2,000 uniform points, 200 corners and the centre of each box.

For the word test, each of the 21,619 candidate words, the alphabetic whole words of the vocabulary that are not function words, takes the free word's place in the real sentence, and blocks 1 to 3 give its point.

\tool runs with $N = 400$ and the defaults of Appendix~\ref{app:exp:setup}, on CPU with 8 threads per job.

\end{document}